\documentclass{article}

\usepackage[final]{neurips_2024}

\usepackage[utf8]{inputenc} % allow utf-8 input
\usepackage[T1]{fontenc}    % use 8-bit T1 fonts
\usepackage{hyperref}       % hyperlinks
\usepackage{url}            % simple URL typesetting
\usepackage{booktabs}       % professional-quality tables
\usepackage{amsfonts}       % blackboard math symbols
\usepackage{nicefrac}       % compact symbols for 1/2, etc.
\usepackage{microtype}      % microtypography
\usepackage{xcolor}         % colors
\usepackage{times}
\usepackage{latexsym}
\usepackage{amsmath}
\usepackage{amsfonts}
\usepackage{graphicx}
\usepackage{pifont}
\usepackage{graphicx}
\usepackage[T1]{fontenc}
\usepackage{afterpage}
\usepackage[utf8]{inputenc}
\usepackage{multirow}
\usepackage{microtype}
\usepackage{booktabs}
\usepackage{graphicx}
\usepackage{subcaption}
\usepackage{wrapfig}
\usepackage{lipsum}

\title{How do Large Language Models Handle Multilingualism?}

\author{
  \hspace{-0.38cm}Yiran Zhao$^{1,2}$\footnotemark[2] \quad Wenxuan Zhang$^{2,3}$\footnotemark[3] \quad Guizhen Chen$^{2,4}$\footnotemark[4]  \quad Kenji Kawaguchi$^{1}$
   \quad Lidong Bing$^{2,3}$ \\
  $^1$ National University of Singapore \quad  $^2$ DAMO Academy, Alibaba Group, Singapore \\ 
  $^3$ Hupan Lab, 310023, Hangzhou, China \quad $^4$ Nanyang Technological University, Singapore \\
}

\begin{document}

\maketitle
\renewcommand{\thefootnote}{\fnsymbol{footnote}}
\footnotetext[2]{This work was done during the internship of Yiran Zhao at Alibaba DAMO Academy.}
\footnotetext[3]{Wenxuan Zhang is the corresponding author: \url{isakzhang@gmail.com}}
\footnotetext[4]{Guizhen Chen is under the Joint Ph.D. Program between DAMO Academy and NTU.}
\renewcommand{\thefootnote}{\arabic{footnote}}

\begin{abstract}
Large language models (LLMs) have demonstrated impressive capabilities across diverse languages. This study explores how LLMs handle multilingualism. Based on observed language ratio shifts among layers and the relationships between network structures and certain capabilities, we hypothesize the LLM's multilingual workflow (\texttt{MWork}): LLMs initially understand the query, converting multilingual inputs into English for task-solving. In the intermediate layers, they employ English for reasoning and incorporate multilingual knowledge with self-attention and feed-forward structures, respectively. In the final layers, LLMs generate responses aligned with the original language of the query. 
To verify \texttt{MWork}, we introduce Parallel Language-specific Neuron Detection (\texttt{PLND}) to identify activated neurons for inputs in different languages without any labeled data. Using \texttt{PLND}, we validate \texttt{MWork} through extensive experiments involving the deactivation of language-specific neurons across various layers and structures. 
Moreover, \texttt{MWork} allows fine-tuning of language-specific neurons with a small dataset, enhancing multilingual abilities in a specific language without compromising others. This approach results in an average improvement of $3.6\%$ for high-resource languages and $2.3\%$ for low-resource languages across all tasks with just $400$ documents.\footnote{Our code is available at \url{https://github.com/DAMO-NLP-SG/multilingual_analysis}}

\end{abstract}

\section{Introduction}\label{sec:introduction}
% \bing{It seems not needed to say a lot of good words for LLMs, because everybody knows this well, can keep space for more important content.}
Recent advancements in large language models (LLMs) ~\citep{openai2023gpt4, touvron2023llama, team2023gemini} have dramatically transformed the field of natural language processing (NLP). 
% integrating seamlessly into daily and professional uses.
Thanks to the extensive pretraining on massive corpora mixed with different languages, these models demonstrate remarkable capabilities in understanding and generating text across multiple languages~\citep{huang-etal-2023-languages, m3exam, zhao2024llama}. 
Despite these advancements,
%in understanding and generating text across multiple language~\citep{huang-etal-2023-languages, zhu2023extrapolating, m3exam, zhao2024llama}, 
the intricate mechanism of their multilingual processing behavior remains largely unclear, which leads to an important research question: \textit{How do large language models handle multilingualism?} 

To understand the working mechanism of LLMs, existing studies mainly focus on the relationship between model architectures and certain capabilities, with some investigating reasoning abilities with self-attention layers~\citep{hou-etal-2023-towards, stolfo-etal-2023-mechanistic, friedman2023interpretability}, and others interpreting feed-forward layers as key-value memories for storing factual knowledge~\citep{geva2021transformer, dai2022knowledge, meng2022locating}. However, these works solely center on English and neglect the multilingual features of LLMs in their interpretations.

\begin{figure}[t]
\centering
\begin{subfigure}[b]{0.45\textwidth}
\includegraphics[width=\textwidth]{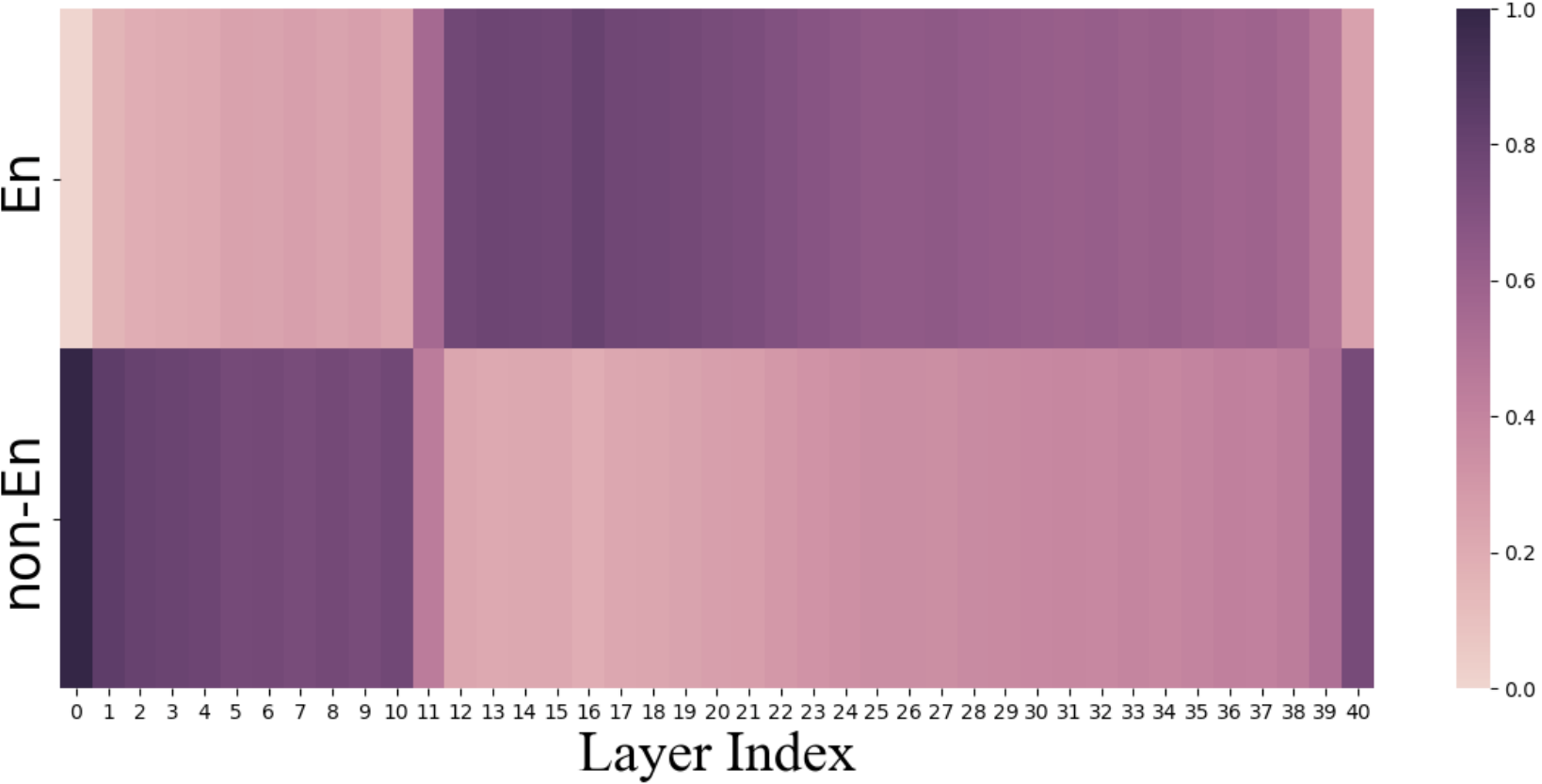}
\caption{Vicuna-13b-v1.5}
\label{fig:dis_vicuna}
\end{subfigure}
\hfill
\begin{subfigure}[b]{0.45\textwidth}
\includegraphics[width=\textwidth]{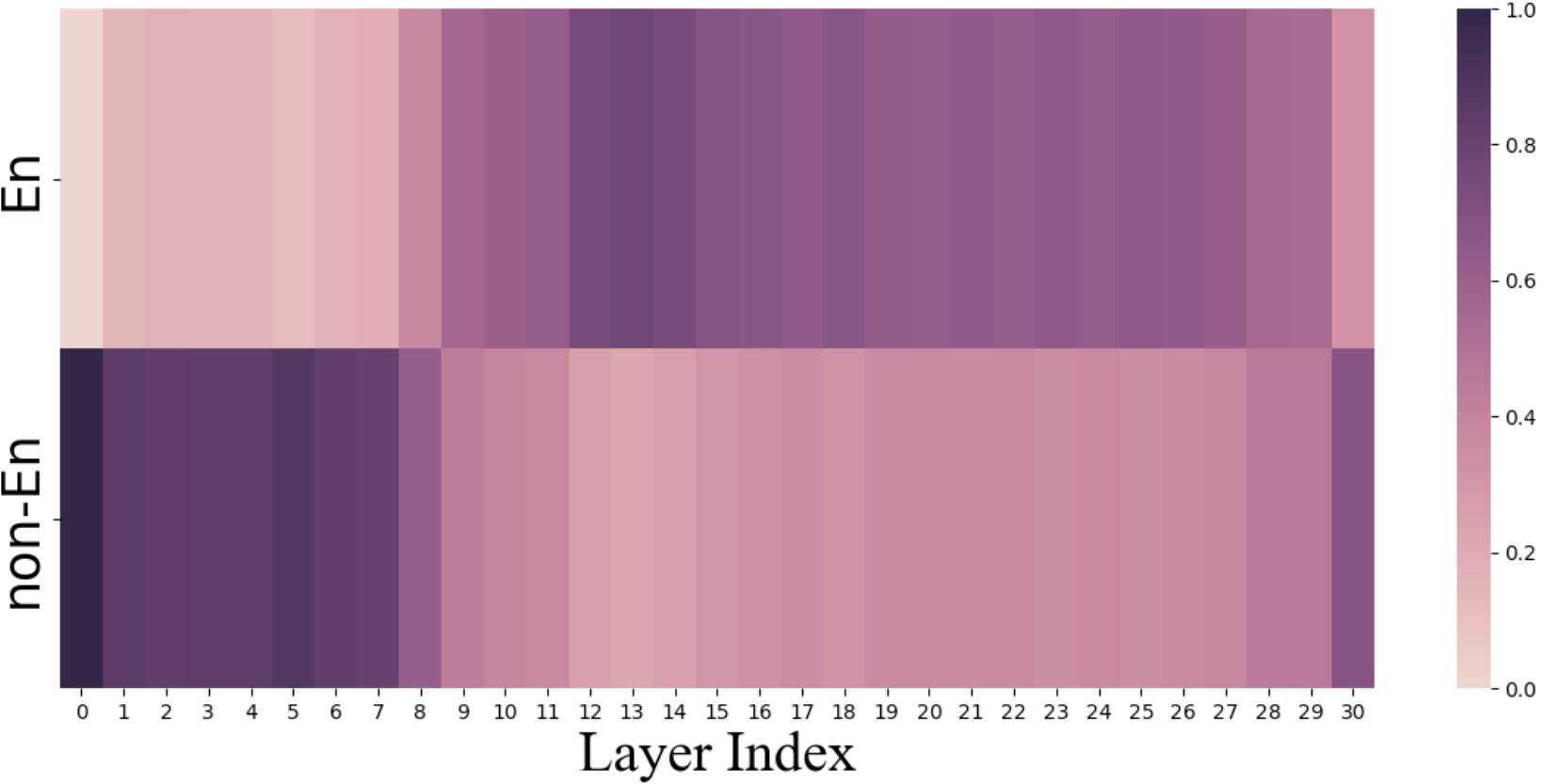}
\caption{BLOOMZ-7b1}
\label{fig:dis_bloomz}
\end{subfigure}
\caption{Ratio of English and non-English tokens among layers given non-English queries.}
\label{fig:dis}
\vspace{-0.3cm}
\end{figure}

% \begin{figure}[t]
% \centering
% \includegraphics[width=0.8\linewidth]{Figures/neurons.pdf}
% \caption{Our proposed multilingual workflow of LLMs.}
% \label{fig:all}
% \end{figure}

To gain an initial understanding of the multilingual mechanism of LLMs, we test LLMs with various non-English queries and decode the hidden embeddings of each layer to tokens within the LLM's vocabulary.
%, utilizing the same decoder as the last layer. 
Subsequently, we classify these decoded tokens %hidden representations 
into either English or non-English, and analyze the ratio. Figure~\ref{fig:dis} illustrates the ratio of English and non-English tokens for each layer of two LLMs. 
%We can observe that non-English queries are initially represented by non-English embeddings as expected. However, as the queries are being processed through the middle layers of the model, the representations become English-centric. In the final few layers, there is a reversion to a predominance of non-English embeddings, corresponding to the non-English queries. 
% \bing{``as the initial instructions are posted in non-English languages'' -$>$ ``in corresponding to the non-English instructions''}
We observe that non-English queries initially generate non-English embeddings as expected. However, as queries progress through the middle layers, the representations surprisingly become English-centric. In the final layers, there is a reversion to predominantly non-English embeddings, matching the non-English queries. 
%In summary, this shift from non-English to English-centric representations and back to non-English describes the transformation workflow within the LLMs for multilingual queries.
% \isak{In summary, this shift from non-English to English-centric representations (in the middle layers) and back to non-English (in the final layers) describes the transformation workflow within the LLMs for multilingual queries.}
%This shift from non-English to English-centric and back to non-English summarizes the transformation workflow within LLMs for multilingual queries.

% \bing{The above paragraph is very important, but not well written. }

\begin{wrapfigure}{r}{0.6\textwidth}
\vspace{-0.3cm}
\centering
    \includegraphics[width=0.6\textwidth]{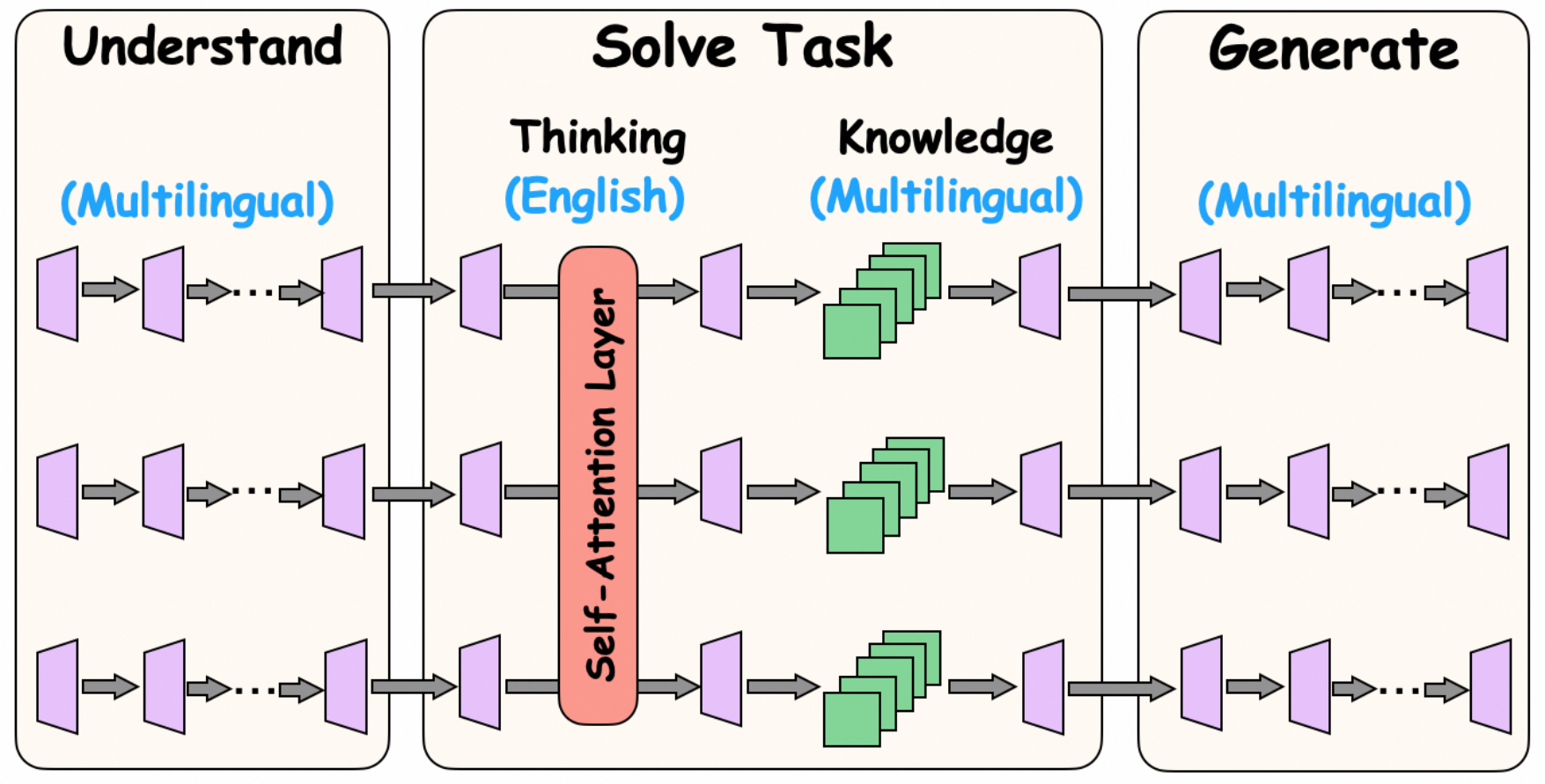}
\caption{Our hypothesized multilingual workflow, \texttt{MWork}, converts multilingual queries to English for reasoning in English and generates responses in the original language, demonstrating a layered processing approach. }
\label{fig:all}
\vspace{-0.3cm}
\end{wrapfigure}

%Motivated by the transformation observed above, we put forth a hypothesis regarding a three-stage multilingual workflow. This workflow entails \textit{understanding} queries in non-English and converting them to English, performing \textit{task-solving} in English, and \textit{generating} outputs that revert back to non-English. \isak{English and non-English what? these are adj.}
%These stages align with the transformation process from non-English to English and subsequently back to non-English. \isak{this sentence just repeats the first one}
%Furthermore, building upon the works of \citept{hou-etal-2023-towards, stolfo-etal-2023-mechanistic, friedman2023interpretability} who investigate the reasoning capabilities of LLMs through self-attention layers, and drawing inspiration from \citept{geva2021transformer, dai-etal-2022-knowledge, meng2022locating} who consider the feed-forward layer as a mechanism for storing factual knowledge in key-value memories, 
%Furthermore, building upon previous studies that interpret self-attention layers for reasoning and feed-forward layers for storing factual knowledge,
%we propose to further decouple the task-solving stage into thinking by self-attention structure and extracting multilingual knowledge with the feed-forward structure. 

Motivated by the observed transformation above, we hypothesize a three-stage multilingual workflow: \textit{understanding}, \textit{task-solving}, and \textit{generating}. This involves understanding the original non-English queries and interpreting them in English, solving tasks in English, and reverting outputs back to the original language. Furthermore, building upon previous studies that link self-attention structures to reasoning and feed-forward structures to factual knowledge storage \citep{hou-etal-2023-towards, geva2021transformer}, we further decouple the task-solving stage into reasoning with self-attention structures and extracting multilingual knowledge with feed-forward structures.
Therefore, our hypothesized Multilingual Workflow (\texttt{MWork}) illustrated in Figure \ref{fig:all} outlines the three operational stages of LLMs in processing multilingual queries: Initially, LLMs \textit{understand} queries by converting diverse linguistic features into a unified representation. In the \textit{task-solving} phase, LLMs reason in English and incorporate multilingual knowledge to obtain factual content, using self-attention and feed-forward structures, respectively. Finally, models \textit{generate} responses in the original language as the original query. 
%\textcolor{blue}{Considering all the above, we propose the LLM's multilingual workflow hypothesis (\texttt{MWork}), as illustrated} in Figure \ref{fig:all} that conceptualizes the operational stages of LLMs when processing multilingual \textcolor{blue}{queries}. 
%In the first several layers, LLMs \textit{understand} the user \textcolor{blue}{query} and convert the diverse linguistic features into a unified representation. 
%Transitioning into the \textit{task-solving} phase, LLMs engage in solving the tasks by thinking in English and incorporating multilingual knowledge to obtain factual contents, leveraging the self-attention and feed-forward structures, respectively.
%Finally, models \textit{generate} responses that align with the original language of the query.

% \bing{The last sentence ``Such a workflow..'' is terrible, it just forces reviewers to buy your conclusion, which is very dangerous. First, your Figure 1 cannot perfectly explain your MWork, how do you know the knowledge part is multilingual from Figure 1? Second, what are the previous studies? you have said no previous studies on the multilingualism features of LLMs.}

% \bing{in the above paragraph, we have instruction, input, and query, they are referring to the same thing.}

% \bing{Note that reviewers may not buy your MWork hypothesis up to here, but in the next paragraph, you start to talk about ``the subsequent question'', which makes reviewers feel even more badly.}

To verify the proposed \texttt{MWork}, we could extract language-specific parameters,  selectively deactivate them within different structures, and observe their corresponding effects, thereby assessing the functionality of corresponding structures and validating our hypothesis. 
To identify the parameters to be activated, we develop a novel approach called Parallel Language-specific Neuron Detection (\texttt{PLND}). Unlike existing methods that rely on fine-tuning\citep{frankle2018lottery, zhang2023fine}, labeled data \citep{tang2024language, liu2024unraveling}, or parallel corpora \citep{libovicky2020language, tanti2021language, zhang2024unveiling} to detect activated parameters, \texttt{PLND} measures the significance of individual neurons with respect to the input in both attention and feed-forward structures without any labeled data or parameter adjustments.
% To tackle this question, it is crucial to pinpoint which parameters are activated by the input, even without explicit labels of certain tasks. \textcolor{blue}{Some studies employ fine-tuning \citep{frankle2018lottery, aghajanyan2021intrinsic, zhang2023fine}, while others rely on parallel corpora \citep{libovicky2020language, tanti2021language, zhang2024unveiling} to detect language-specific parameters. However, these approaches either require labeled data or introduce parameter changes, resulting in increased interpretational uncertainty. 
% In addition, works focus on LLMs solely focus on the feed-forward layer \citep{tang2024language, liu2024unraveling}. 
% Moreover, which requires labeled data and introduces parameter changes to LLMs, leading to increased uncertainty in their interpretation. Moreover, \citept{libovicky2020language, tanti2021language, zhang2024unveiling} rely on parallel corpus to investigate language representations. In addition, works focus on LLMs such as \citept{tang2024language, liu2024unraveling} solely focus on the feed-forward layer. 
% We} develop a novel approach called Parallel Language-specific Neuron Detection (\texttt{PLND}), which effectively measures the significance of individual neurons \textcolor{blue}{with respect to} the input provided. 
% \isak{still too long and will make readers lost, shorten it to just 1-2 sentences}
Using \texttt{PLND}, we identify language-specific neurons by inputting a free text corpus of that language and isolating consistently activated neurons. We find that by deactivating language-specific neurons which account for only $0.13\%$ of all neurons, LLMs' performance on a multilingual summarization task could drop by $99\%$.

We then extensively verify the hypothesized \texttt{MWork} framework using the proposed \texttt{PLND} method. Employing various benchmark tasks, including XQuAD~\citep{artetxe2020cross} for understanding, MGSM~\citep{shi2022language} for reasoning, X-CSQA~\citep{lin2021common} for knowledge extraction, and XLSum for generation~\citep{hasan2021xl}, we selectively deactivate language-specific neurons in each component and verify the functionality of the component by observing a significant decline in performance on the corresponding task. 
For example, when deactivating the language-specific neurons in the understanding layer, the performance on the multilingual understanding task XQuAD remains stable in English, while experiencing a decrease of $14\%$ in non-English languages. Other tasks exhibit similar pattern when deactivating corresponding neurons. 
More importantly, with the verified \texttt{MWork} framework, enhancing the multilingual capabilities of LLMs can thus be achieved through the fine-tuning of language-specific neurons for certain capabilities. With a remarkable reduction in the training corpus size to a mere few hundred documents, this fine-tuning procedure enhances the multilingual capabilities of LLMs for both high-resource and low-resource languages by an average of $3.6\%$ and $2.3\%$ across all tasks, respectively. Notably, even without an English training corpus, there is a noticeable improvement in English performance, as the enhancement of language-specific neurons yields greater accuracy in enhancing specific languages, while simultaneously ensuring a clear division of parameters among different languages.
% \isak{add a half sentence explaining why, such as ``as such enhancement is more accurate and doesn't affect...'' However, in the previous sentence, you said ``XQuad remains stable in English'' is this a conflict?}.
% . We fine-tune Mistral-7b-base on 400 documents, resulting in average performance increases of $3.4\%$ on MGSM, $4.4\%$ on XQuAD, $4.3\%$ on X-CSQA, and $2.3\%$ on XLSum for high-resource languages. Low-resource languages (Thai, Vietnamese, Arabic, and Swahili) benefit with an average improvement of $2.2\%$ across four tasks. }
%\isak{two key parts: 1. we extensively verify it; 2. this can be used for enhancement -- make them more clear}
In summary, the verified \texttt{MWork} reveals how LLMs handle multilingual tasks and offers an effective approach for conducting language-specific enhancements %demonstrating that by implementing it and fine-tuning language-specific neurons in specific structures using a small dataset, we can enhance the multilingual capabilities of a specific language 
without compromising performance in other languages.
% \isak{this sentence is relatively simple, you can further edit it. the main idea is to put a summary sentence here}
%The hypothesized \texttt{MWork} workflow is consistent with the observations made above and aligns with findings from previous studies. This suggests the potential for LLMs to transition between languages, enabling them to effectively comprehend, reason, and respond to multilingual tasks.

\section{Parallel Language-specific Neuron Detection (\texttt{PLND})}\label{sec:2}

To verify the hypothesized workflow, we propose \texttt{PLND} that effectively detects language-specific neurons without relying on any labeled data. In essence, \texttt{PLND} identifies neurons crucial for handling individual documents, with language-specific neurons being those that consistently show high importance when processing documents in a particular language.
% In this section, we introduce a neuron detection method called \texttt{PLND} for \textcolor{blue}{effectively detecting language-specific neurons.}
% \isak{To verify the hypothesized workflow, we describe our proposed neuron detection method named \texttt{PLND} for effectively detecting language-specific neurons in this section.}
%\isak{maybe even add 1-2 more sentences to elaborate}

\subsection{Sequential Neuron Detection}
We define a neuron as a single row or column of a parameter matrix of a language model.
To identify neurons responsible for a specific language, it is crucial to discern the significance of a neuron with respect to the inference of a given input.
%Note that in the following sections, a neuron is defined as a single row or column of the parameter matrix with a size of $hidden\_size$. 
Specifically, when processing the input $c$ in the model, we denote the hidden embedding before the $i$-th layer in Transformer~\citep{vaswani2017attention} as $h_i$, and the hidden embedding after the $i$-th layer as $h_{i+1} = T_i(h_i)$, where $T_i$ represents the parameters of the $i$-th layer. For a specific neuron within the $i$-th layer, denoted as $N^{(i)}$, either located in the attention or feed-forward network, we quantify its importance in processing the input $c$ by measuring the difference in the hidden embedding after the $i$-th layer, i.e., $h_{i+1}$, when $N^{(i)}$ is activated or deactivated. Formally, the impact of neuron $N^{(i)}$ for input $c$ is defined as
\begin{equation}
\text{Imp}(N^{(i)}|c) = \|T_{i}\backslash N^{(i)}(h_i) - T_{i}(h_i)\|_2,
\label{equ:imp}
\end{equation}
where $T_{i}\backslash N^{(i)}(\cdot)$ denotes deactivating $N^{(i)}$ in $T_i$, i.e., setting all parameters of the neuron $N^{(i)}$ to zero. With a set of $n$ corpus in a specific language, denoted as $\mathcal{C} = \{c_1,\cdots,c_l,\cdots,c_n\}$, we calculate the importance of each neuron in each layer to each corpus. Furthermore, we can obtain language-specific neurons that are important to all corpus in that language, i.e., 
\begin{equation}
\{N^{(i)} \mid \text{Imp}(N^{(i)}|c_l)\geq \epsilon, \forall c_l\in\mathcal{C}\},
\end{equation} where $\epsilon$ is the pre-defined threshold.

\subsection{Parallel Neuron Detection}

The sequential neuron detection requires traversal of all neurons and inputs sequentially and thus is time-consuming. To address this, we further propose a parallel algorithm for accelerating the process.

\paragraph{Feed-Forward Network (FFN)}

In the latest open-source models, when processing input $c$, the feed-forward network in a certain layer is defined as
\begin{equation}
\text{FFN}(x) = \Big(\text{SiLU}\big(W_{gate}(x)\big)\cdot W_{up}(x)\Big)W_{down},
\end{equation}
where $x\in\mathbb{R}^{l\times d_{model}}$ is the embedding fed into the FFN, $W_{gate},W_{up}\in\mathbb{R}^{d_{model}\times d_{inter}}$\footnote{$W(\cdot)$ represents the linear matrix product of the input $x$ and the parameter $W$, i.e., $W(x):=xW$.}, $W_{down}\in\mathbb{R}^{d_{inter}\times d_{model}}$. The calculation of the importance of the $k$-th neuron in $W_{up}$, when processing the input $c$, as presented in Equation \ref{equ:imp}, can be equivalently transformed to
\begin{equation}
\text{Imp}(W_{up}[:,k]|c) =  \|\hat{\text{FFN}}(x) - \text{FFN}(x) \|_2   = \Big\| \big(h_{\text{ffn}}\cdot \texttt{Mask}[k]\big) W_{down}(x)\Big\|_2,
\end{equation}
where $h_{\text{ffn}}\in d_{inter}$ represents the embedding before $W_{down}$, and $\texttt{Mask}[k]\in d_{inter}$ is a vector with the $k$-th element equal to $1$ and the rest equal to $0$. To calculate $\text{Imp}(W_{up}[:,k]|c)$ for $k\in d_{inter}$ parallelly, we introduce a diagonal mask matrix of size $(d_{inter}, d_{inter})$, denoted as \texttt{Mask}. Therefore,
\begin{equation}
\begin{aligned}
& \text{Imp}(W_{up}|c) =  \|( h_{\text{ffn}}\cdot \texttt{Mask}) W_{down}(x)\|_2.\hspace{-0.5cm}
\end{aligned}\label{equ:imp_ffn}
\end{equation}
Furthermore, we observe that deactivating the $k$-th neuron of $W_{down}$ is equivalent to deactivating the $k$-th neuron in $W_{up}$, as they both result in $h_{\text{ffn}}[k] = 0$. Hence, we can also derive $\text{Imp}(W_{down}|c)$ by employing Equation (\ref{equ:imp_ffn}).

\paragraph{Self-Attention Network}

When processing input $c$, the self-attention network in a certain layer is 
\begin{equation}
\begin{aligned}
\text{Attention}(x) = \text{Softmax}\big(\frac{W_Q(x)W_K^T(x)}{\sqrt{d}}\big)W_V(x),
\end{aligned}
\end{equation}
where $W_Q, W_K, W_V\in\mathbb{R}^{d_{model}\times d_{mid}}$. 
% $W_K\in\mathbb{R}^{d_{model}\times d_{mid}}$, $W_V\in\mathbb{R}^{d_{model}\times d_{mid}}$.
\footnote{In some models like Vicuna and Mistral, $d_{model}=d_{mid}$, but we use different notations to avoid ambiguity.} Since $W_V(x)$ is not in the non-linear softmax calculation, we can calculate $\text{Imp}(W_{V}|c)$ by applying Equation (\ref{equ:imp_ffn}). For $W_Q$, we obtain $\text{Imp}(W_{Q}[:,k]|c)$ by deactivating its $k$-th neuron, specifically, $\hat{W}_{Q}\leftarrow W_{Q}[:,k] = 0$. Firstly, we calculate the difference in attention weight before and after deactivation, prior to scaling and softmax,
\begin{equation}
\begin{aligned}
\Delta_k (x) & = W_Q(x)W_K^T(x) - \hat{W}_Q(x)W_K^T(x) = W_{Q}(x)[:,k]W_K(x)[k,:] \in\mathbb{R}^{l\times l}. \\
\end{aligned}
\end{equation}
Next, as the changes in attention exhibit a positive correlation with the changes in the output of this layer, the importance of $W_Q[:,k]$ in processing $c$, as defined in Equation \ref{equ:imp}, can be approximated as
\begin{equation}
\begin{aligned}
\text{Imp}(W_Q[:,k]|c) & \approx \|\hat{\text{attention}}(x) - \text{attention}(x)\|_2 \\
&\approx \Big\|\text{softmax}\big(\frac{W_Q(x)W_K^T(x)-\Delta_k(x)}{\sqrt{d}}\big) - \text{softmax}\big(\frac{W_Q(x)W_K^T(x)}{\sqrt{d}}\big)\Big\|_2.\\
\end{aligned}\label{equ:imp_att}
\end{equation}
This process can also be calculated in parallel, specifically,
\begin{equation}
\begin{aligned}
\Delta (x) =  &\; W_Q(x)W_K^T(x) - \hat{W}_Q(x)W_K^T(x)  \\
 = &\; W_{Q}(x).resize(l,1,d_{mid})\times W_K(x).resize(1,l,d_{mid}) \in\mathbb{R}^{l\times l\times d_{mid}}. \\
\end{aligned}
\end{equation}
Therefore, the importance of $W_Q$ in processing input $c$ is calculated by
\begin{equation}
\begin{aligned}
\text{Imp}(W_Q|c)
&\approx \Big\|\text{softmax}\big(\frac{W_Q(x)W_K^T(x)-\Delta(x)}{\sqrt{d}}\big) - \text{softmax}\big(\frac{W_Q(x)W_K^T(x)}{\sqrt{d}}\big)\Big\|_2.\\
\end{aligned}\label{equ:final}
\end{equation}
Similarly, since $W_K$ is symmetrical to $W_Q$, $\text{Imp}(W_{K}|c)$ can be calculated in the same way.

\subsection{Detection of Language-Specific Neurons}

We then apply \texttt{PLND} to selected languages and models to validate its effectiveness in detecting language-specific neurons and to further investigate the relationships between languages.

\paragraph{Experimental Setup.} 
% We test on \textit{Mistral-7b-Instruct-v0.2}~\citep{jiang2023mistral}, which is fine-tuned on \textit{Mistral-7B-v0.2}. For simplicity, we abbreviate it as Mistral hereafter. 
We test two open-source models that perform well on multilingual tasks, including \textit{Vicuna-7b-v1.5}\footnote{We do not directly utilize Llama2-chat as it does not follow multilingual instructions, consistently responding in English regardless of the language of the query.}~\citep{vicuna2023} and \textit{Mistral-7b-Instruct-v0.2}~\citep{jiang2023mistral}. For simplicity, we abbreviate them as Vicuna and Mistral hereafter to represent the two models respectively.
We select the text summarization task with the  XLSum~\citep{hasan2021xl} dataset as the reference task to evaluate multilingual performance as it requires the model to comprehend the input text and generate a coherent fragment. 
%Failing this task indicates a loss of language ability to generate coherent fragments.
%, while inability to complete other tasks suggests a potential reduction in reasoning ability but retains understanding and effective fragment generation.
We adopt $4$ high-resource languages including French (Fr), Chinese (Zh), Spanish (Es), and Russian (Ru), as their initial performance on those languages is already quite reasonable for observing the multilingual processing mechanism.
%levels are sufficiently high to allow for the observation of significant drops after deactivation. 
Furthermore, we utilize OSCAR~\citep{caswell2020language} corpus which contains web crawling texts for each language to compile a language-specific corpus without task-specific considerations. More details are presented in Appendix \ref{appen:corpus}. 

\paragraph{Existence of Language-Specific Neurons}
Using \texttt{PLND}, we feed a corpus in a specific language to LLMs and identify neurons that are consistently activated, which are responsible for processing queries in that language.
To ascertain whether these neurons are genuinely language-specific,
we assess the performance of LLMs in corresponding languages when these neurons are deactivated versus when the same number of randomly sampled neurons are deactivated. 
% We utilize the summarization task with XLSum dataset~\citep{hasan2021xl} to showcase the capabilities of LLMs, aiming to demonstrate that the resulting output can be devoid of meaning. 

\begin{table*}[t]
\caption{Multilingual performance on XLSum when deactivating language-specific neurons (``Lang-Spec'') and an equivalent number of randomly selected neurons (``Random'').}
  \centering
\footnotesize
  \scalebox{0.95}{
  \begin{tabular}{ll|c|c|c|c|l}
    \toprule
   \normalsize{\textbf{Model}} & \normalsize{\textbf{Method}}  & \normalsize{\textbf{Fr}} & \normalsize{\textbf{Zh}} & \normalsize{\textbf{Es}} & \normalsize{\textbf{Ru}} & \normalsize{\textbf{Avg.}} \\
    \midrule
  \multirow{3}*{{\textbf{Vicuna}}}  & Original & $14.2$ & $61.1$ & $10.4$ & $20.8$ & $26.6$ \\
  & Deactivate Random & $14.1 $ & $61.6$ & $10.4$ & $20.8$ & $26.7$ \\
 & Deactivate Lang-Spec & $\mathbf{0.83}$ & $\mathbf{0.00}$ &$\mathbf{0.24}$ & $\mathbf{0.42}$ & $\mathbf{0.37}$   \\\midrule
\multirow{3}*{{\textbf{Mistral}}}   & Original & $15.2$ & $56.4$ & $10.6$ & $21.0$ & $25.8$ \\
& Deactivate Random & $15.4$ & $55.9$ & $10.2$ & $21.2$ & $25.7$ \\
 & Deactivate Lang-Spec & $\mathbf{0.21}$ & $\mathbf{0.39}$ & $\mathbf{0.15}$ & $\mathbf{0.07}$ & $\mathbf{0.21}$ \\
    \bottomrule  
    \end{tabular}}\label{table:lang_neu}
\vspace{-0.3cm}
\end{table*}

Table \ref{table:lang_neu} demonstrates the decline of multilingual capabilities when deactivating language-specific neurons. Although just deactivating around $0.13\%$ neurons, LLMs lose their multilingual capabilities and fail to generate meaningful content.
In contrast, deactivating the same number of randomly selected neurons does not yield any difference.
Therefore, the detected neurons are language-specific and related to handling corresponding multilingual inputs. 
%\textcolor{blue}{However, \citept{tang2024language} requires deactivating $3\%$ of neurons to achieve a $50\%$ reduction in multilingual performance, while \citept{liu2024unraveling} extract $10\%$ parameters in FFN layer as language-specific. Therefore, \texttt{PLND} is the most effective language-specific neutron detection method, which empowers us to verify \texttt{MWork}.}

\subsection{Analysis of Language-Specific Neurons}

We further investigate the degree of overlap among their language-specific neurons. Our findings reveal that in both Mistral and Vicuna, English shows limited overlap with other languages, indicating many language-specific neurons, while languages within the same family, such as Spanish, French, and English, demonstrate more overlap. More details are illustrated in Appendix \ref{appen:interneuron}.
%\isak{consider whether adding the comparisons => they are not directly comaprable}

In addition, we examine two more types of multilingual LLMs, including BLOOMZ~\citep{muennighoff2023crosslingual}, a \textit{hyper-multilingual} LLM claiming to support 46 languages, and Chinese Llama \citep{cui2023efficient}, a \textit{bilingual} LLM focusing on English and Chinese. We find that language-specific neurons in BLOOMZ follow patterns similar to Mistral and Vicuna. However, in Chinese LLama, Chinese dominates as the primary language for reasoning and knowledge extraction across all languages, with notably absent language-specific neurons. Details are shown in Appendix \ref{appen:more}.

Given the certain overlap ratio of language-specific neurons from other languages with those of English, as illustrated in the first column of Figure \ref{fig:lang_influ} and Figure \ref{fig:lang_influen_bloom}, we conduct supplementary experiments to demonstrate that these neurons are not language-agnostic neurons crucial for general comprehension and logical reasoning~\citep{liang2024multilingual, tang2024language}. Instead, these overlapping neurons represent only a subset of language-specific neurons, while the language-agnostic neurons responsible for essential understanding and reasoning are those not identified as language-specific. Further elaboration and detailed results are presented in Appendix \ref{appen_agno}.

\section{Multilingual Workflow (\texttt{MWork}) of LLMs}\label{sec:3}

\subsection{\texttt{MWork}}

\begin{wrapfigure}{R}{0.5\textwidth}
\vspace{-0.3cm}
\centering
    \includegraphics[width=0.5\textwidth]{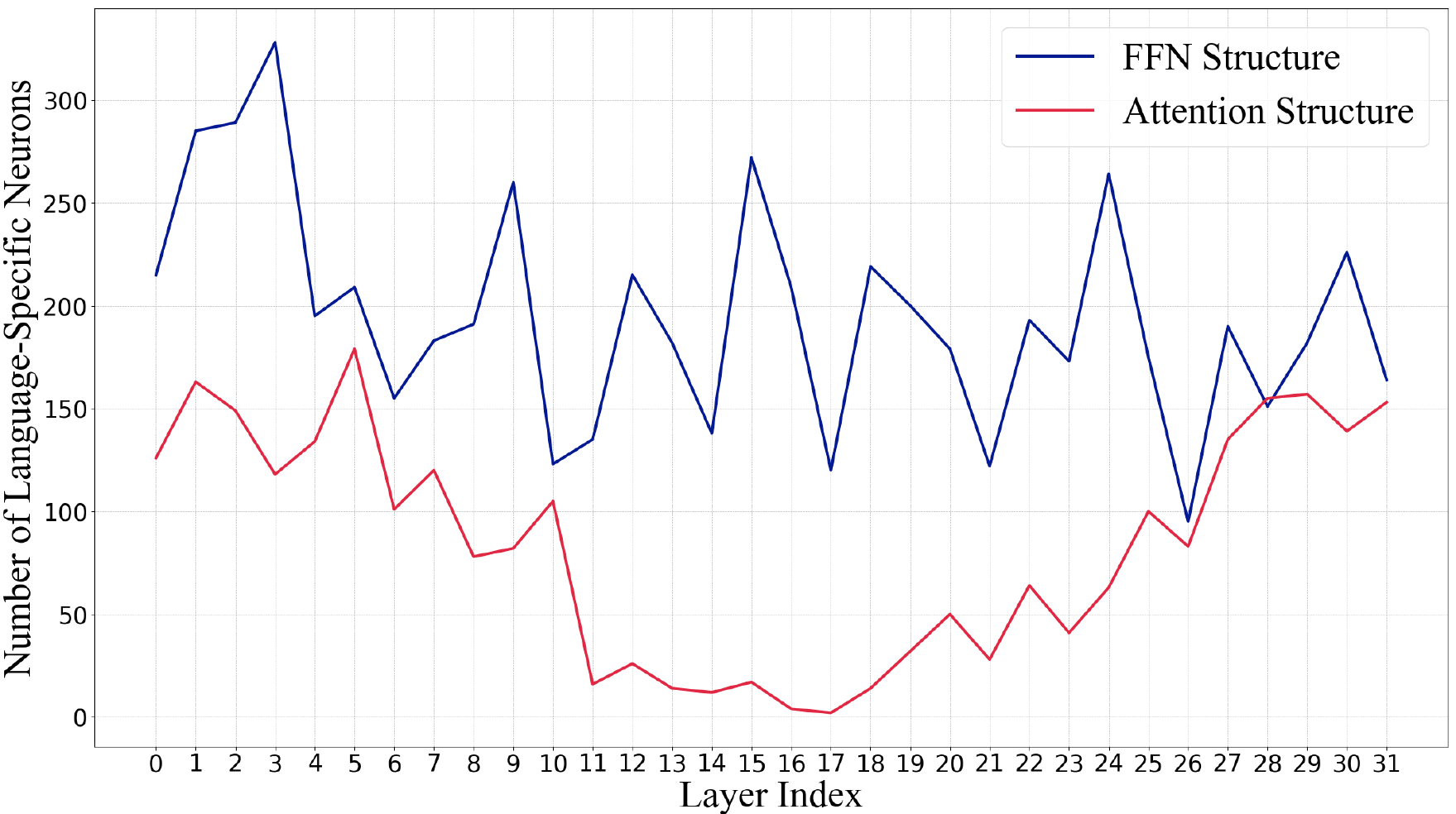}
\caption{Number of language-specific neurons when processing multilingual queries.}
\label{fig:neuron_sta}
\vspace{-0.3cm}
\end{wrapfigure}

By classifying the hidden representations of each layer in LLMs into English or non-English (as shown in Figure \ref{fig:dis}), we can observe the shift from non-English to English-centric, and back to non-English with the progression through the layers. This motivates us to hypothesize a three-stage multilingual workflow: \textit{understanding} the original non-English queries and interpreting them in English, \textit{task-solving} in English, and \textit{generating} back to the original language.
Nevertheless, the presence of certain non-English tokens during the English-centric task-solving stage inspires us to further investigate this stage.

%As mentioned earlier, we hypothesize a three-stage multilingual workflow: \textit{understanding} the original non-English queries and interpreting them in English, \textit{task-solving} in English, and \textit{generating} back to the original language, as motivated by Figure \ref{fig:dis}.
% Motivated by the ratio between English and non-English tokens in LLMs when handling multilingual queries (Figure \ref{fig:dis}), we hypothesize a three-stage multilingual workflow: \textit{understanding} the original non-English queries and interpreting them in English, \textit{task-solving} in English, and \textit{generating} back to the original language. 
%Nevertheless, the presence of non-English tokens during the English-centric task-solving stage allows for a greater potential to finely divide the task-solving layer. We extract language-specific neurons from attention and feed-forward structures by \texttt{PLND} while processing $50$ multilingual queries and average the number of activated language-specific neurons across queries, as illustrated in Figure \ref{fig:neuron_sta}.
% With \texttt{PLND}, language-specific neurons are extracted from attention and feed-forward structures during the processing of a single multilingual query, as depicted in Figure \ref{fig:neuron_sta}. 

With the proposed \texttt{PLND} method, we extract language-specific neurons from attention and feed-forward structures when processing various multilingual queries. We plot the average number of activated language-specific neurons of Mistral when processing each query in Figure \ref{fig:neuron_sta}.
Notably, the number of language-specific neurons decreases within the self-attention structure in the task-solving layer but remains consistent across the layers of the feed-forward structure. This decline implies a reliance on the English language for reasoning while extracting multilingual knowledge to support query processing, which is also consistent with \citep{geva2021transformer}'s interpretation of the feed-forward structure as key-value memories for knowledge extraction. 
Therefore, we further decompose the task-solving layer into two parts: \textit{reasoning} in English and \textit{extracting knowledge} in a multilingual context. 

Considering the above insights, we propose the \texttt{MWork} hypothesis for explaining LLM's multilingual workflow: LLMs first \textit{understand} user input by unifying diverse linguistic features. They then engage in the \textit{task-solving} phase, employing English for reasoning and leveraging multilingual knowledge through self-attention and feed-forward structures, respectively. Finally, the models \textit{generate} responses aligned with the query's original language.

% \textcolor{blue}{Fortunately, the presence of language-specific neurons, as verified above, greatly facilitates the verification of \texttt{MWork}. It empowers us to investigate the intricate functions of neurons and systematically deactivate language-specific neurons throughout various layers and components. Such exploration allows us to examine the impact of these neurons on a wide range of tasks in a fine-grained manner.}

\subsection{Verification Experiment Setup}

To verify \texttt{MWork}, we selectively deactivate language-specific neurons from each component. Then its functionality can be verified if this deactivation results in minimal impact on English performance while exhibiting a notable decline in multilingual performance for the corresponding task.
\paragraph{Dataset}
To comprehensively understand how LLMs work with different abilities, we employ four kinds of tasks including MGSM~\citep{shi2022language} for reasoning task, XQuAD~\citep{artetxe2020cross} for understanding task, X-CSQA~\citep{lin2021common} for knowledge question answering task, and XLSum~\citep{hasan2021xl} for generation task. Detailed information regarding these datasets and the testing prompts can be found in Appendix \ref{appen:prompt}. We adopt $6$ languages including English (En), German (De), French (Fr), Chinese (Zh), Spanish (Es), and Russian (Ru), as their initial performance on those languages is already quite reasonable for observing the multilingual processing mechanism. For XLSum, we randomly sample $500$ data points from the whole test set for each language taking into consideration its long inference time, while for other tasks, we employ the entire test set. We evaluate the vanilla performance of Vicuna and Mistral on these datasets for later comparison as presented in Appendix \ref{appen:orig}. For reasoning, understanding, and knowledge question answering tasks, we adopt accuracy as the metric. As for the generation tasks, we adopt ROUGE-L as the metric.

\paragraph{Deactivation Strategy} We primarily consider two aspects when selecting the deactivation settings: (1) language-specific neurons versus randomly chosen neurons, and (2) the position of neurons, which encompasses four structures. Note that for a fair comparison, we ensure the numbers of deactivated neurons in all settings are the same. More detailed settings are explained from Section \ref{sec:under} to Section \ref{sec:gen}. For the concrete numbers of different layers, we tune hyperparameters by XQuAD in Chinese. Details are explained in Appendix \ref{appen:hyper}. 
% \isak{can change the title to ``Evaluation Strategy'', then elaborate how such deactivation can validate the hypothesis}

\begin{table*}[t]
\caption{Results of the \textbf{understanding} task, where `\ding{55}' indicates that chosen neurons in the corresponding layer are deactivated, and `\ding{51}' signifies they are activated.  $\Delta$ is defined as the difference between the reduction in performance in English, denoted as $\Delta_{\text{Eng}}$, and the reduction in performance in non-English languages, denoted as $\Delta_{\text{n-Eng}}$.}
  \centering
\footnotesize
  \scalebox{0.92}{
  \begin{tabular}{cccccc|ccccc}
    \toprule
   \multirow{2}*{\textbf{\normalsize{Model}}}  & \multicolumn{5}{c}{\textbf{\normalsize{Deactivating Method}}} \vline
     & \multicolumn{5}{c}{\textbf{\normalsize{Performance}}} \\
  & Under & S-ATTN & S-FFN & Gen & Neuron & Eng & n-Eng & $\Delta_{\text{Eng}}$ & $\Delta_{\text{n-Eng}}$ & $\Delta$ $\uparrow$   \\\midrule
  \multirow{5}*{{\normalsize{Vicuna}}} 
& \ding{55} & \ding{51} & \ding{51} & \ding{51} & Random & \colorbox{white}{$57.8$} & $53.9$ & $+0.3$ & $-0.1$ & $+0.4$ \\
& \ding{55} & \ding{55} & \ding{55} & \ding{55} & Random & \colorbox{white}{$57.9$} & $54.2$ & $+0.4$ & $+0.3$ & $+0.1$ \\
& \ding{51} & \ding{55} & \ding{55} & \ding{51} & Lang-Spec & \colorbox{white}{$40.9$} & $38.6$ & $-15.9$ & $-15.3$ & $-0.6$   \\
& \ding{51} & \ding{51} & \ding{51} & \ding{55} & Lang-Spec & \colorbox{white}{$57.9$} & $52.8$ & $-0.4$ & $-1.1$ & $+0.7$ \\
& \ding{55} & \ding{51} & \ding{51} & \ding{51} & Lang-Spec & $56.5$ & $46.0$ & \colorbox{lightgray}{$-0.5$} & \colorbox{lightgray}{$-7.9$} & \colorbox{lightgray}{$+7.4$} \\\midrule
  \multirow{5}*{\normalsize{Mistral}}
& \ding{55} & \ding{51} & \ding{51} & \ding{51} & Random & \colorbox{white}{$58.1$} & $55.5$ & $+1.0$ & $-0.2$ & $+1.2$ \\
& \ding{55} & \ding{55} & \ding{55} & \ding{55} & Random & \colorbox{white}{$57.6$} & $55.5$ & $+0.5$ & $-0.2$ & $+0.7$ \\
& \ding{51} & \ding{55} & \ding{55} & \ding{51} & Lang-Spec &\colorbox{white}{$53.2$} & $47.0$ & $-3.9$ & $-8.7$ & $+4.8$   \\
& \ding{51} & \ding{51} & \ding{51} & \ding{55} & Lang-Spec & \colorbox{white}{$56.4$} & $54.6$ & $-0.7$ & $-1.0$ & $+0.3$ \\
& \ding{55} & \ding{51} & \ding{51} & \ding{51} & Lang-Spec &$56.2$ & $48.3$ & \colorbox{lightgray}{$-0.9$} & \colorbox{lightgray}{$-7.4$} & \colorbox{lightgray}{$+6.5$} \\
    \bottomrule  
    \end{tabular}}
\label{table:result_under_all}
\vspace{-0.1cm}
\end{table*}

\paragraph{Notations} Tables \ref{table:result_under_all} to \ref{table:result_gen_all} present the results of deactivating certain neurons, where ``Under'' denotes the understanding layers, ``S-ATTN'' and ``S-FFN'' correspond to the self-attention and the feed-forward structures within the task-solving layers respectively, ``Gen'' refers to the generation layers. The term ``Random'' is used to describe deactivating randomly chosen neurons, whereas ``Lang-Spec'' refers to the deactivation of language-specific neurons.
We also present the gap between the original performance (as shown in Table \ref{table:vanilla}) and performance after deactivation (as shown in Table \ref{table:result_under} to Table \ref{table:result_gen}) for English ($\Delta_{\text{Eng}}$) and averaged non-English languages ($\Delta_{\text{n-Eng}}$), respectively. A single metric $\Delta$ is then introduced as $\Delta_{\text{Eng}} - \Delta_{\text{n-Eng}}$, where a high value indicates such deactivation operation does not bring much impact to the English performance but lead to performance drop in non-English. Therefore, this provides a direct single indicator that the deactivated neurons are language-specific and hold a significant responsibility in executing the corresponding task.
% \isak{add one sentence to convey the idea that: therefore a higher value of delta means it verifies our corresponding assumption}

% \isak{where is the division of three parts - understanding, task-solving, generation layers}

\subsection{Verify the Understanding Stage in \texttt{MWork}}\label{sec:under}

\paragraph{Deactivating Method} Table \ref{table:result_under_all} shows the results of the understanding task following the deactivation of five distinct sets of neurons: (i) neurons randomly selected from the understanding layers; (ii) neurons randomly chosen across all layers; (iii) language-specific neurons within the task-solving layers; (iv) language-specific neurons in the generation layers; (v) language-specific neurons in the understanding layers.
% For a fair comparison, we ensure the numbers of deactivated neurons in all settings are the same. 
As mentioned above, in order to verify the functionality of the understanding layer (setting v), we compare it with deactivating other types of layers, specifically setting iii for the task-solving layer and setting iv for the generation layer. Full results are listed in Appendix \ref{appen:detail}.

\paragraph{Findings} We find that by deactivating randomly sampled neurons, no matter in the understanding layer or all layers, the performance of LLMs in both English and non-English languages is almost unaffected compared to other deactivating methods. Note that in some cases, deactivating randomly sampled neurons may even increase the performance because irrelevant neurons are removed, which also aligns with the finding from~\citep{sharma2023truth}. When assessing the differential impact on English and non-English language performance after the deactivation, specifically the difference calculated as $\Delta_{\text{Eng}} - \Delta_{\text{n-Eng}}$, it is evident that the deactivation of random neurons within the understanding layer amplifies this effect. This observation lends partial support to the hypothesized role of the understanding layer in language processing.

Furthermore, we find that deactivating language-specific neurons in the understanding layer influences the performance in English a little while significantly decreasing the performance in non-English languages. When deactivating language-specific neurons in the task-solving layer, both English and non-English languages are significantly reduced while deactivating language-specific neurons in the generation layer influences a little for both English and non-English languages. Therefore, we prove that the first several layers are responsible for understanding because deactivated neurons just disable LLMs on the NLU task in non-English languages. Furthermore, disabling language-specific neurons in the task-solving layer shows that LLMs rely on English, as performance drops across all languages.

\subsection{Verify the Reasoning Structure in \texttt{MWork}}

\begin{table*}[t]
\caption{Results of the \textbf{reasoning} task. Disabling all language-specific neurons, except for those involved in self-attention structure within the task-solving layer, greatly reduces performance.}
  \centering
\footnotesize
  \scalebox{0.92}{
  \begin{tabular}{cccccc|ccccc}
    \toprule
\multirow{2}*{\textbf{\normalsize{Model}}}  & \multicolumn{5}{c}{\textbf{\normalsize{Deactivating Method}}} \vline
     & \multicolumn{5}{c}{\textbf{\normalsize{Performance}}} \\
  & Under & S-ATTN & S-FFN & Gen & Neuron & Eng & n-Eng & $\Delta_{\text{Eng}}$ & $\Delta_{\text{n-Eng}}$ & $\Delta$ $\uparrow$   \\\midrule
  \multirow{6}*{{\normalsize{Vicuna}}} 
& \ding{51} & \ding{55} & \ding{51} & \ding{51} & Random & \colorbox{white}{$20.0$} & $11.3$ & $-0.4$ & $-1.8$ & $+1.4$ \\
& \ding{51} & \ding{55} & \ding{55} & \ding{51} & Random & \colorbox{white}{$18.4$} & $12.2$ & $-2.0$ & $-1.0$ & $-1.0$ \\
& \ding{55} & \ding{55} & \ding{55} & \ding{55} & Random &\colorbox{white}{$19.6$} & $12.5$ & $-0.8$ & $-0.7$ & $-0.1$ \\
& \ding{51} & \ding{55} & \ding{55} & \ding{51} & Lang-Spec & \colorbox{white}{$7.2$} & $3.4$ & $-13.2$ & $-9.8$ & $-3.4$   \\
& \ding{55} & \ding{51} & \ding{51} & \ding{55} & Lang-Spec & \colorbox{white}{$18.1$} & $8.3$ & $-2.3$ & $-4.9$ & $+2.6$ \\
& \ding{55} & \ding{51} & \ding{55} & \ding{55} & Lang-Spec &  \colorbox{white}{$19.0$} & $7.8$ & \colorbox{lightgray}{$-1.4$} & \colorbox{lightgray}{$-5.4$} & \colorbox{lightgray}{$+4.0$} \\\midrule
  \multirow{6}*{{\normalsize{Mistral}}}
& \ding{51} & \ding{55} & \ding{51} & \ding{51}  & Random &\colorbox{white}{$40.8$} & $23.4$ & $-5.2$ & $-2.9$ & $-2.3$ \\
& \ding{51} & \ding{55} & \ding{55} & \ding{51}  & Random & \colorbox{white}{$39.2$} & $24.0$ & $-6.8$ & $-2.3$ & $-4.5$ \\
& \ding{55} & \ding{55} & \ding{55} & \ding{55} & Random & \colorbox{white}{$45.2$} & $26.8$ & $-0.8$ & $+0.5$ & $-1.3$ \\
& \ding{51} & \ding{55} & \ding{55} & \ding{51} & Lang-Spec & \colorbox{white}{$38.2$} & $18.4$ & $-7.8$ & $-7.9$ & $+0.1$   \\
& \ding{55} & \ding{51} & \ding{51} & \ding{55} & Lang-Spec & \colorbox{white}{$44.0$} & $18.1$ & $-2.0$ & $-8.2$ & $+6.2$ \\
& \ding{55} & \ding{51} & \ding{55} & \ding{55} & Lang-Spec &  \colorbox{white}{$46.2$} & $18.3$ & \colorbox{lightgray}{$+0.2$} & \colorbox{lightgray}{$-8.0$} & \colorbox{lightgray}{$+8.2$} \\
    \bottomrule  
    \end{tabular}}
\label{table:result_reason_all}
\end{table*}

\paragraph{Deactivating Method}
Table \ref{table:result_reason_all} shows the result of the reasoning task, where we deactivate $6$ sets of neurons. We adhere to the previous logic of selecting deactivation settings, with the exception that we do not conduct an independent experiment on deactivating neurons in the understanding layer, as its functionality has already been verified. Details are listed in Appendix \ref{appen:detail}.

\paragraph{Findings} We find that deactivating randomly sampled neurons in task-solving layers disables the capabilities of LLMs in reasoning to a greater extent than deactivating randomly sampled neurons in all layers, which verifies the function of the task-solving layer. Furthermore, comparing three deactivating language-specific neuron methods, we find that deactivating the task-solving layer decreases performance in both English and non-English. On the contrary, when we only deactivate language-specific neurons not in the task-solving layer, non-English is influenced more seriously than English. Moreover, eliminating interference from the feed-forward layer achieves better results, which verifies the function of attention structure in the task-solving layer.

\subsection{Verify the Knowledge Extraction Structure in \texttt{MWork}}

\paragraph{Deactivating Method}
Table \ref{table:result_know_all} shows the result of the knowledge question answering task, where we deactivate $5$ sets of neurons. Similarly, we exclude the deactivation of neurons in layers that have already been verified and instead concentrate on the self-attention structure and feed-forward structure in the task-solving layer. Details are listed in Appendix \ref{appen:detail}. 

\begin{table*}[t]
\caption{Results of the \textbf{knowledge} question answering task. The highest performance reduction difference ($\Delta$) is achieved by disabling all language-specific neurons in the feed-forward structure within the task-solving layer.}
  \centering
\footnotesize
  \scalebox{0.92}{
  \begin{tabular}{cccccc|ccccc}
    \toprule
\multirow{2}*{\textbf{\normalsize{Model}}}  & \multicolumn{5}{c}{\textbf{\normalsize{Deactivating Method}}} \vline
     & \multicolumn{5}{c}{\textbf{\normalsize{Performance}}} \\
  & Under & S-ATTN & S-FFN & Gen & Neuron & Eng & n-Eng & $\Delta_{\text{Eng}}$ & $\Delta_{\text{n-Eng}}$ & $\Delta$ $\uparrow$   \\\midrule
  \multirow{5}*{{\normalsize{Vicuna}}} 
& \ding{51} & \ding{51} & \ding{55} & \ding{51} & Random & \colorbox{white}{$57.5$} & $39.5$ & $-0.3$ & $+0.0$ & $-0.3$ \\
& \ding{51} & \ding{55} & \ding{55} & \ding{51} & Random & \colorbox{white}{$56.0$} & $38.7$ & $-1.8$ & $-0.8$ & $-1.0$ \\
& \ding{55} & \ding{55} & \ding{55} & \ding{55} & Random & \colorbox{white}{$57.7$} & $39.6$ & $-0.1$ & $+0.1$ & $-0.2$ \\
& \ding{51} & \ding{55} & \ding{51} & \ding{51} & Lang-Spec & \colorbox{white}{$33.7$} & $30.3$ & $-24.1$ & $-9.2$ & $-14.9$ \\
& \ding{51} & \ding{51} & \ding{55} & \ding{51} & Lang-Spec & \colorbox{white}{$57.5$} & $37.5$ & \colorbox{lightgray}{$-0.3$} & \colorbox{lightgray}{$-2.0$} & \colorbox{lightgray}{$+1.7$}   \\\midrule
  \multirow{5}*{{\normalsize{Mistral}}}
& \ding{51} & \ding{51} & \ding{55} & \ding{51} & Random &\colorbox{white}{$61.0$} & $37.0$ & $-0.3$ & $-0.5$ & $+0.2$ \\
& \ding{51} & \ding{55} & \ding{55} & \ding{51} & Random & \colorbox{white}{$60.7$} & $36.3$ & $-0.6$ & $-1.2$ & $+0.6$ \\
& \ding{55} & \ding{55} & \ding{55} & \ding{55} & Random &\colorbox{white}{$61.8$} & $37.4$ & $+0.1$ & $-0.1$ & $+0.2$ \\
& \ding{51} & \ding{55} & \ding{51} & \ding{51} & Lang-Spec & \colorbox{white}{$51.2$} & $28.9$ & $-10.1$ & $-8.6$ & $-1.5$   \\
& \ding{51} & \ding{51} & \ding{55} & \ding{51} & Lang-Spec & {$61.2$} & $35.1$ & \colorbox{lightgray}{$-0.1$} & \colorbox{lightgray}{$-2.4$} & \colorbox{lightgray}{$+2.3$} \\
    \bottomrule  
    \end{tabular}}
\label{table:result_know_all}
\end{table*}

\paragraph{Findings} Likewise, 
% the deactivation of randomly selected neurons has a \textcolor{blue}{less} impact compared to language-specific neurons, validating the efficiency of \texttt{PLND} in identifying neurons pertinent to a particular language. 
targeted deactivation of language-specific neurons within the feed-forward structure of the task-solving layer predominantly affects non-English languages. This implies that processing multilingual queries necessitates accessing the multilingual information embedded within the relevant structures. However, disabling the self-attention structure compromises the ability to solve tasks across all languages.

\subsection{Verify the Generation Structure in \texttt{MWork}}\label{sec:gen}

\paragraph{Deactivating Method}
Table \ref{table:result_gen_all} shows the result of the generation task, where we deactivate $3$ sets of neurons. Since all previous layers have been verified, we solely deactivate neurons in the generation layer and compare them with randomly selected neurons. Details are listed in Appendix \ref{appen:detail}.

\begin{table*}[t]
\caption{Results of the \textbf{generation} task. The highest performance reduction difference ($\Delta$) is achieved by disabling all language-specific neurons in the generation layer.}
  \centering
\footnotesize
  \scalebox{0.92}{
  \begin{tabular}{cccccc|ccccc}
    \toprule
\multirow{2}*{\textbf{\normalsize{Model}}}  & \multicolumn{5}{c}{\textbf{\normalsize{Deactivating Method}}} \vline
     & \multicolumn{5}{c}{\textbf{\normalsize{Performance}}} \\
  & Under & S-ATTN & S-FFN & Gen & Neuron & Eng & n-Eng & $\Delta_{\text{Eng}}$ & $\Delta_{\text{n-Eng}}$ & $\Delta$ $\uparrow$   \\\midrule
  \multirow{3}*{{\normalsize{Vicuna}}} 
& \ding{51} & \ding{51} & \ding{51} & \ding{55} & Random &  \colorbox{white}{$13.2$} & $26.8$ & $+0.1$ & $+0.1$ & $+0.0$ \\
& \ding{55} & \ding{55} & \ding{55} & \ding{55} & Random & \colorbox{white}{$13.0$} & $26.7$ & $-0.1$ & $+0.0$ & $-0.1$ \\
& \ding{51} & \ding{51} & \ding{51} & \ding{55} & Lang-Spec &  \colorbox{white}{$13.1$} & $25.7$ & \colorbox{lightgray}{$+0.0$} & \colorbox{lightgray}{$-1.1$} & \colorbox{lightgray}{$+1.1$} \\\midrule
  \multirow{3}*{{\normalsize{Mistral}}}
& \ding{51} & \ding{51} & \ding{51} & \ding{55} & Random & \colorbox{white}{$13.6$} & $25.9$ & $+0.1$ & $+0.1$ & $+0.0$ \\
& \ding{55} & \ding{55} & \ding{55} & \ding{55} & Random & \colorbox{white}{$13.6$} & $25.7$ & $+0.1$ & $-0.2$ & $+0.3$ \\
& \ding{51} & \ding{51} & \ding{51} & \ding{55} & Lang-Spec &\colorbox{white}{$13.8$} & $24.3$ & \colorbox{lightgray}{$+0.3$} & \colorbox{lightgray}{$-1.5$} & \colorbox{lightgray}{$+1.8$} \\
    \bottomrule  
    \end{tabular}}
\label{table:result_gen_all}
\end{table*}

\paragraph{Findings} Similar to other tasks, the disabling of language-specific neurons within the generation layer diminishes their capacity to generate content in the respective languages. By selectively deactivating neurons that are not associated with English, we do not completely eliminate the models' multilingual generation abilities. However, as demonstrated in Table \ref{table:lang_neu}, the complete deactivation of all language-specific neurons results in the total loss of the LLMs' multilingual generation capabilities.

\begin{figure}[t]
  \begin{minipage}{0.54\textwidth}
    \centering
    \includegraphics[width=\linewidth]{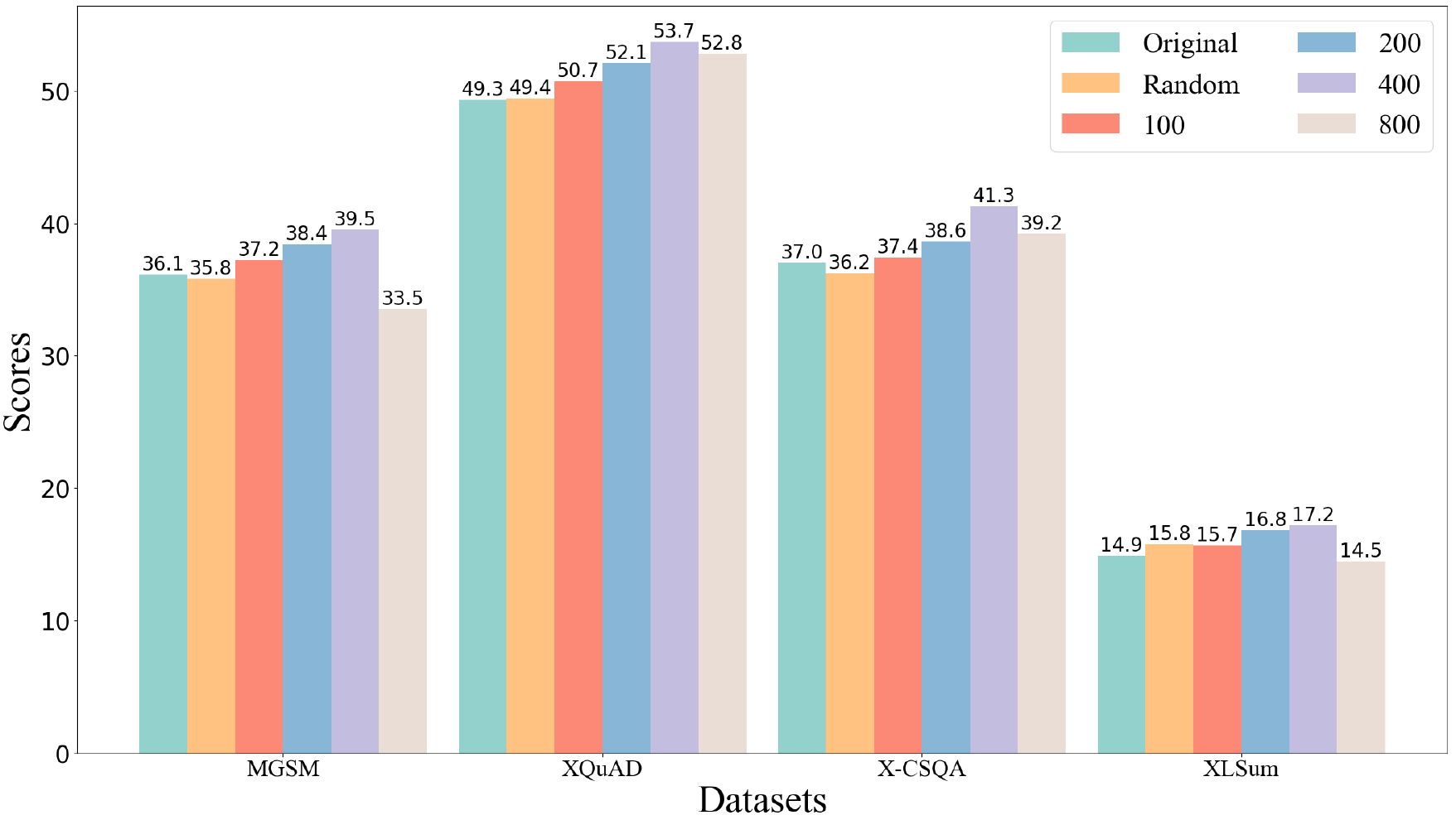}
    \captionof{figure}{Enhancement results on high-resource languages, while the number is average among languages.}
    \label{fig:enhance}
  \end{minipage}
  \hfill
  \begin{minipage}{0.43\textwidth}
    \captionof{table}{Enhancement is achieved by fine-tuning Mistral-7b-v0.1 model utilizing $400$ documents from each language correspondingly. The results are averaged across four tasks. Performance on English (``En'') is obtained by averaging the results from four fine-tuned models.}
    \centering
    \footnotesize
    \setlength{\tabcolsep}{3.0pt}
    \scalebox{0.97}{
      \begin{tabular}{l|ccccc}
        \toprule
        \textbf{\normalsize{Method}}
        & \textbf{\normalsize{En}}  & \textbf{\normalsize{Vi}} & \textbf{\normalsize{Th}} & \textbf{\normalsize{Ar}} & \textbf{\normalsize{Sw}}  \\
        \midrule
        Original & \colorbox{white}{$41.1$} & $32.7$ & $25.6$ & $21.7$ & $15.1$  \\
        Random &  \colorbox{white}{$40.8$} & $32.7$ & $25.2$ & $21.2$ & $15.1$  \\
        Lang-Spec &  \colorbox{white}{$\mathbf{44.6}$} & $\mathbf{34.9}$ & $\mathbf{28.5}$ & $\mathbf{23.4}$ & $\mathbf{16.9}$ \\
        \bottomrule  
      \end{tabular}}
    \label{table:result_enhance}
  \end{minipage}
\vspace{-0.2cm}
\end{figure}

\section{Multilingual Enhancement with \texttt{MWork}}\label{sec:4}
We have verified \texttt{MWork} for explaining the multilingual working mechanism of LLMs in the above section via deactivating certain neurons.
While opposite to employing deactivation, we can also enhance their multilingual ability, especially the understanding and generating ability, by fine-tuning these language-specific neurons. With language-specific neurons comprising only around $0.1\%$ of all parameters, the need for training documents to improve multilingual capabilities can be significantly reduced to just a few hundred. Additionally, fine-tuning only the language-specific neurons for a particular language does not impact performance in other languages, allowing us to enhance specific languages while preserving performance in others.

% It is important to note that our enhancements are focused on augmenting the model's capabilities in understanding and generation only; we do not extend its reasoning capabilities or broaden its knowledge base as it may require more specific data preparation. 
\paragraph{\texttt{MWork} helps with enhancing multilingual ability by hundreds of documents.} 
% \isak{the summary is not good, where is our method} 
We employ \textit{Mistral-7b-v0.1} for enhancement to eliminate the interference of instruction fine-tuning, and select causal language modeling as our training task. We create a dataset comprising $\{100, 200, 400, 800\}$ randomly selected documents for each language, extracted from the Wikipedia corpus~\citep{wikidump}.
% \isak{no need to talk about num of doc here since this is exp details. In this paragraph, mainly explain the idea of ``MWork can also be used for multilingual''} 
Figure \ref{fig:enhance} shows the results of enhancement on high-resource languages (De, Fr, Zh, Es, Ru). The numbers represent the sizes of the training corpus when fine-tuning language-specific neurons, while "Random" represents the fine-tuning of an equivalent number of randomly chosen neurons using a corpus of $400$. Our findings reveal that fine-tuning with a few hundred documents yields significant performance improvements on multilingual tasks: $3.4\%$ on MGSM, $4.4\%$ on XQuAD, $4.3\%$ on X-CSQA, and $2.3\%$ on XLSum. Moreover, English performance is enhanced by an average of $3.7\%$ across all tasks. These results further confirm the effectiveness of \texttt{MWork} in interpreting structure functionality for LLM's multilingual query handling, offering precise and independent methods for multilingual enhancement.
% \isak{similar issue as the title => where is ``our method'', emphasize why such improvement can be achieved} 
When fine-tuning with 800 documents, the performance deteriorates compared to using 400 documents. This drop can be attributed to the incorporation of additional knowledge, which disrupts the original knowledge distribution and leads to overfitting of the model to Wikipedia. This can be addressed by mixing data from more sources such as textbooks or websites.
% \isak{the last sentence should stop at a ``positive ending'' not just explaining. for example, explain what happened, give the reason, then explain how this can be solved} \isak{also, the current explanation is not good, it shows the disadvantages of our method => the knowledge can be easily disturbed with our method}

In addition, we verify the effectiveness of such enhancement method on low-resource languages, given that low-resource performance is relatively low with the original model. We select four languages including Vietnamese (Vi), Thai (Th), Arabic (Ar), and Swahili (Sw), covering languages with both latin and non-latin scripts and having corresponding testing set in our considered benchmarks. The model was then evaluated on four benchmarks, and the result shown in Table \ref{table:result_enhance} is the average scores among tasks. 
It is evident that the fine-tuning method using language-specific neurons enhances the model's multilingual performance in low-resource languages by an average of $2.2\%$. 
% \isak{then what is the difference with last paragraph? just ``low-resource also get improved''? Two options: 1. find any particular surprising improvement of certain tasks so you can sell the idea ``low-resource can be improved A LOT'' 2. say something special about low-resource language. At least say something like ``although low-resource performance is relatively low with the original model, it can be improved'' (though this is weak)}
Notably, the improvement of $3.5\%$ in English performance is observed even without an English training corpus, indicating the effectiveness of the distinct language responsibilities assigned to neurons. 

\section{Related Work}

In the era of LLMs, numerous studies have been conducted to develop multilingual benchmarks~\citep{m3exam}, enhance multilingual performance without parameter adjustments through translation~\citep{liang2023machine, huang-etal-2023-languages}, aligning representations~\citep{nguyen2023enhancing, salesky2023pixel}, prompting~\citep{li2023enhancing, tanwar2023multilingual}. Furthermore, certain works focus on improving multilingual abilities for a single task via cross-lingual transfer~\citep{kim2017cross, lin2019choosing, pfeiffer2020mad, zhao2024adamergex}, while others aim to enhance multilingual proficiency by continuous training in one language to obtain mono-lingual LLMs \citep{cui2023efficient}, or in multiple domain languages to obtain domain-lingual LLMs~\citep{nguyen2023seallms}. Additionally, some works achieve multilingual LLMs by training from scratch~\citep{muennighoff2023crosslingual}. However, these studies are limited to specific task types or require substantial training corpora due to a lack of comprehensive understanding of the multilingual mechanisms of LLMs.

% \isak{the first sentence does not mention multilingual} These efforts underscore the importance and complexity of enabling LLMs to operate effectively across multiple languages.

Conventional interpretability research investigates the significance of input features with their corresponding outputs~\citep{vig2019multiscale,hewitt2019designing, qiu2020pre}. In the era of LLMs, one brunch of work includes efforts to understand knowledge storage, with \citep{geva2021transformer} initiating the study of the feed-forward layer as a knowledge base. Subsequent work has furthered this by altering neuron values \citep{dai2022knowledge}, mapping embeddings to words \citep{geva2022transformer}, modifying inputs to recover embeddings \citep{meng2022locating}, and analyzing attention heads \citep{li2023inference}. Another line of research centers on the self-attention layer, examining its connection to reasoning capability \citep{{hou-etal-2023-towards,stolfo-etal-2023-mechanistic,friedman2023interpretability}} by contrasting the reasoning tree based on attention weights.

% \paragraph{Multilingualism} Various studies have been undertaken to construct benchmarks~\citep{m3exam}, enhance performance through translation~\citep{liang2023machine, huang-etal-2023-languages}, aligning representations~\citep{nguyen2023enhancing, salesky2023pixel}, prompting~\citep{li2023enhancing, tanwar2023multilingual}. These efforts underscore the importance and complexity of enabling LLMs to operate effectively across multiple languages.

\section{Conclusion}\label{sec:discussion}
In this work, we examine how LLMs handle multilingualism. The proposed multilingual workflow (\texttt{MWork}) suggests that LLMs initially understand queries by converting multilingual inputs into English, reason in English in intermediate layers while incorporating multilingual knowledge, and generate responses aligned with the original language in the final layers. The validity of \texttt{MWork} is verified using Parallel Language-specific Neuron Detection (\texttt{PLND}), which identifies activated neurons for different languages without labeled data. By detecting language-specific neurons and fine-tuning them with a small training corpus, \texttt{MWork} enhances multilingual abilities in specific languages without compromising others, resulting in significant improvements across tasks.

\clearpage

% \section*{Limitation and Impact Statements}

% Our experiments are mainly conducted on models with a size of approximately 7 billion parameters (i.e., vicuna-7b-v1.5 and mistral-7b-v1.0), primarily due to computational resource constraints. Testing our methods on larger models could potentially yield additional insights, particularly in understanding the scalability of our proposed framework. Furthermore, our exploration into enhancing the multilingual abilities of language models through our framework was preliminary. Expanding our experiments to include a broader array of languages, especially those considered low-resource, could better demonstrate the effectiveness of our proposed framework. Moreover, extending the scope of our experiments to evaluate other capabilities such as reasoning and multilingual knowledge extraction with specific datasets could provide a more comprehensive picture of the potential benefits of our approach.

% Our paper has the potential to significantly enhance the multilingual ability of LLMs. By effectively improving their performance across all languages, it can enable the development of real multilingual models that excel in various applications, promoting better communication and understanding among diverse linguistic communities.

\section*{Acknowledgement}
This work was substantially supported by DAMO Academy through DAMO Academy Research Intern Program. This research is partially supported by the National Research Foundation Singapore under the AI Singapore Programme (AISG Award No: AISG2-TC-2023-010-SGIL) and the Singapore Ministry of Education Academic Research Fund Tier 1 (Award No: T1 251RES2207).

\bibliography{anthology,custom}
\bibliographystyle{acl}

\clearpage

\appendix

\section{English and Non-English Tokens}\label{appen:intro}

We employ \texttt{cld3} package to detect the language of each token in the embeddings of each layer, which is a language detection library based on the Compact Language Detector $3$ model developed by Google. Furthermore, if the detection result is reliable, i.e., $\text{cld3.get\_language(token).is\_reliable} == True$, we adopt the detection results, otherwise the token is categorized as a non-word.

\section{Multilingual Corpus}\label{appen:corpus}

Note that our selection criterion for the number of documents is based on achieving substantial coverage of each language's vocabulary, ensuring that the selected contexts provide a representative sample of the language, as shown in Table \ref{table:corpus}.

\begin{table*}[ht]
\caption{Corpus details across languages are tailored to encompass the majority of each language's vocabulary, where ``corpus size'' indicates the number of contexts selected, ``corpus vocab'' represents the vocabulary coverage within the selected contexts, ``vocab size'' refers to the number of vocabularies of that language.}
  \centering
\footnotesize
  \scalebox{0.88}{
  \begin{tabular}{l|c|c|c|c|c|c}
    \toprule
 \textbf{Language} & En & De & Fr  & Zh & Es   & Ru  \\
    \midrule
 Corpus Size & $180$k & $30$k & $50$k & $20$k & $20$k & $20$k \\\midrule
 Corpus Vocab & $249$k & $154$k & $134$k    & $198$k & $90$k   & $144$k   \\\midrule
 Vocab Size & $273$k & $148$k & $135$k  & $329$k & $93$k & $150$k  \\
    \bottomrule  
    \end{tabular}}
\label{table:corpus}
\vspace{-0.2cm}
\end{table*}

\section{Interrelation of Language-Specific Neurons Across Languages}\label{appen:interneuron}

\begin{figure}[ht]
    \centering
    \begin{subfigure}{0.49\textwidth}
        \centering
        \includegraphics[width=\linewidth]{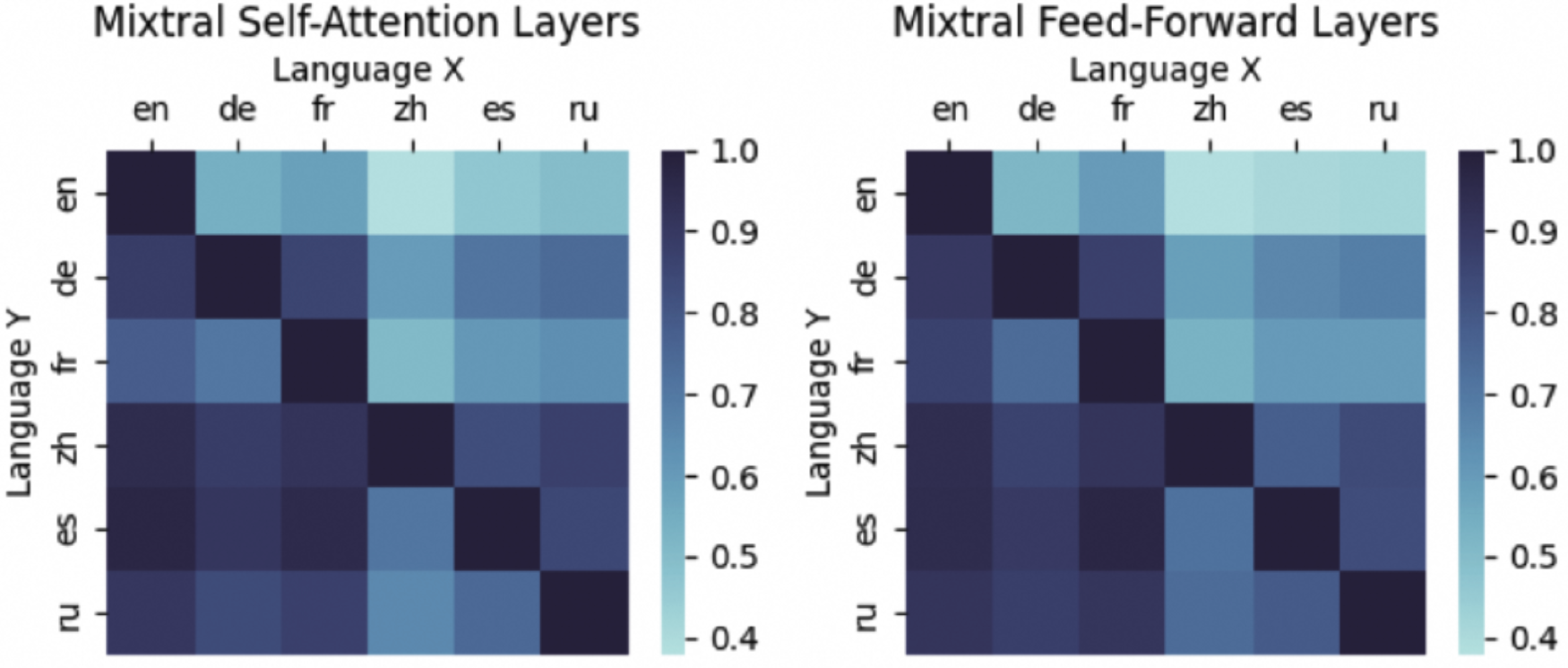}
        \caption{Mistral-7B-Instruct-v0.2.}
        \label{fig:lang_influ_mistral}
    \end{subfigure}\hfill
    \begin{subfigure}{0.49\textwidth}
        \centering
        \includegraphics[width=\linewidth]{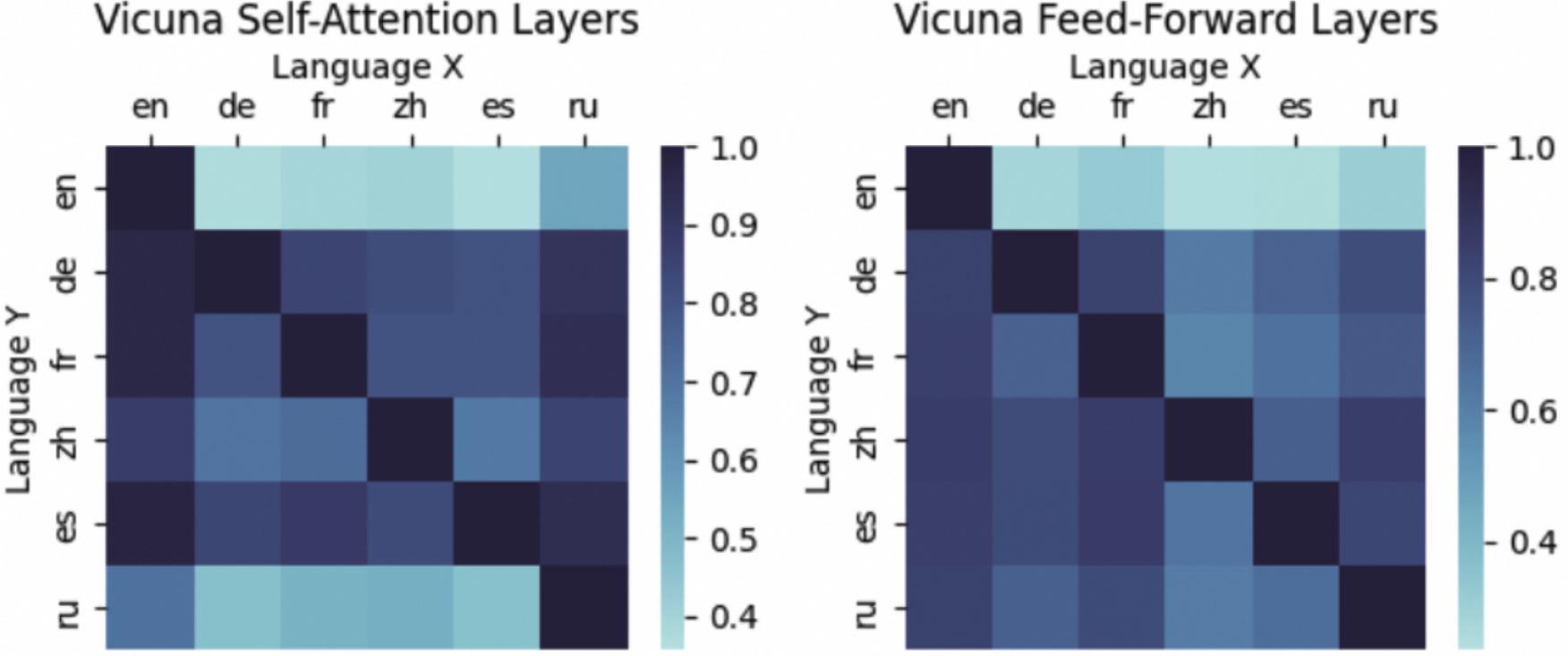}
        \caption{Vicuna-7b-v1.5.}
        \label{fig:lang_influ_vicuna}
    \end{subfigure}
    \caption{Overlapping ratio of language-specific neurons in self-attention and feed-forward structures.}
    \label{fig:lang_influ}
\end{figure}

Using neurons identified by \texttt{PLND}, we investigate the relationships between languages via the degree of overlap among their language-specific neurons, defined as
\begin{equation}
    \text{overlap}(x,y) = \frac{|\mathcal{N}_{x}\cap\mathcal{N}_{y}|}{|\mathcal{N}_{y}|},
\end{equation}
where $\mathcal{N}_{language}$ represents the set of detected language-specific neurons. 
Figure~\ref{fig:lang_influ} shows the neuron overlapping ratio $\text{overlap}(x,y)$ of any two languages in different structures of two models.

We can observe that in both Mistral and Vicuna, the intersection with English from other languages is relatively limited (i.e., the first row of each figure), suggesting that English possesses a predominant number of language-specific neurons. 
Additionally, there is a pronounced tendency for languages belonging to the same family to demonstrate a higher degree of overlap with each other, such as Spanish, French, and English. 
% Moreover, the feed-forward structure typically exhibits a higher degree of consistency in overlap across various languages, due to the shared world knowledge embedded within the neurons that is accessible to multiple languages.

\section{Analysis on Different Multilingual LLMs}\label{appen:more}
We further examine two more types of multilingual LLMs, including BLOOMZ~\citep{muennighoff2023crosslingual}, a \textit{hyper-multilingual} LLM claiming to support 46 languages, and Chinese Llama \citep{cui2023efficient}, a \textit{bilingual} LLM focusing on English and Chinese.

\begin{figure}[ht]
  \centering
  \begin{minipage}[t]{0.48\textwidth} % Adjust the minipage width to fit your content
    \includegraphics[width=\textwidth]{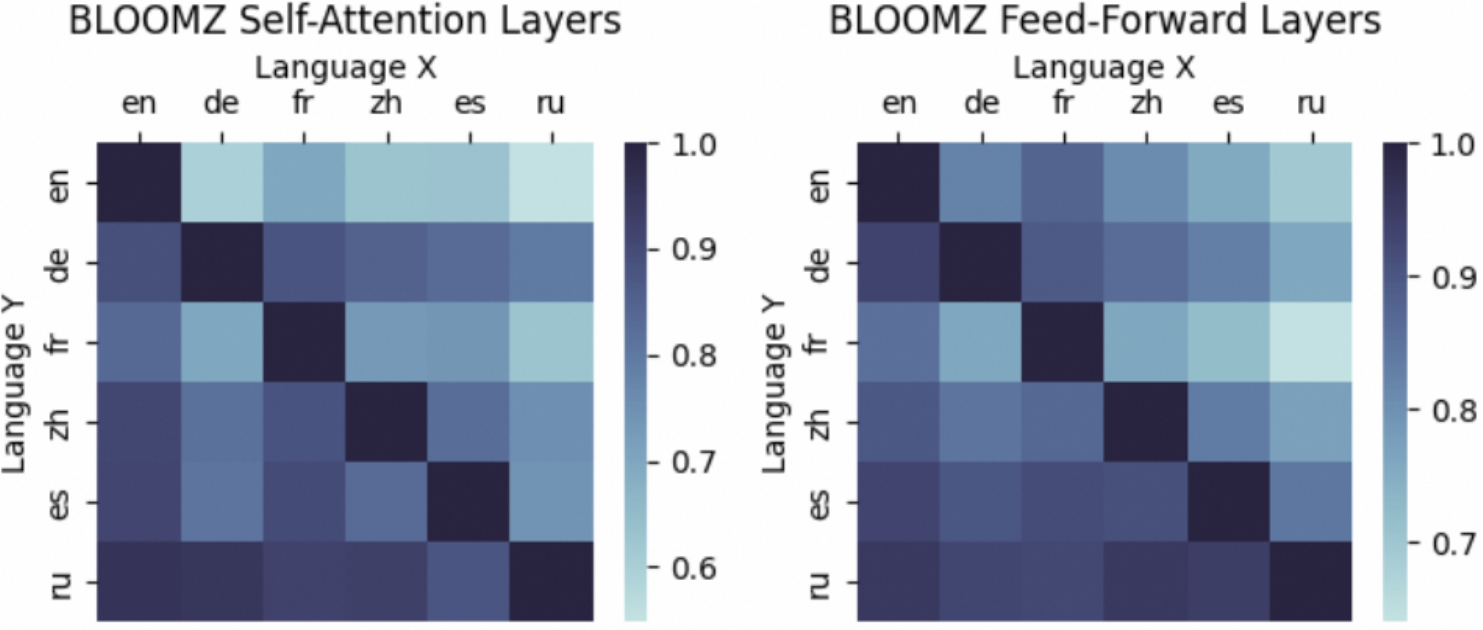}
    \caption{Overlapping ratio of language-specific neurons in BLOOMZ}
    \label{fig:lang_influen_bloom}
  \end{minipage}
  \hfill
  \begin{minipage}[t]{0.48\textwidth}
    \includegraphics[width=\textwidth]{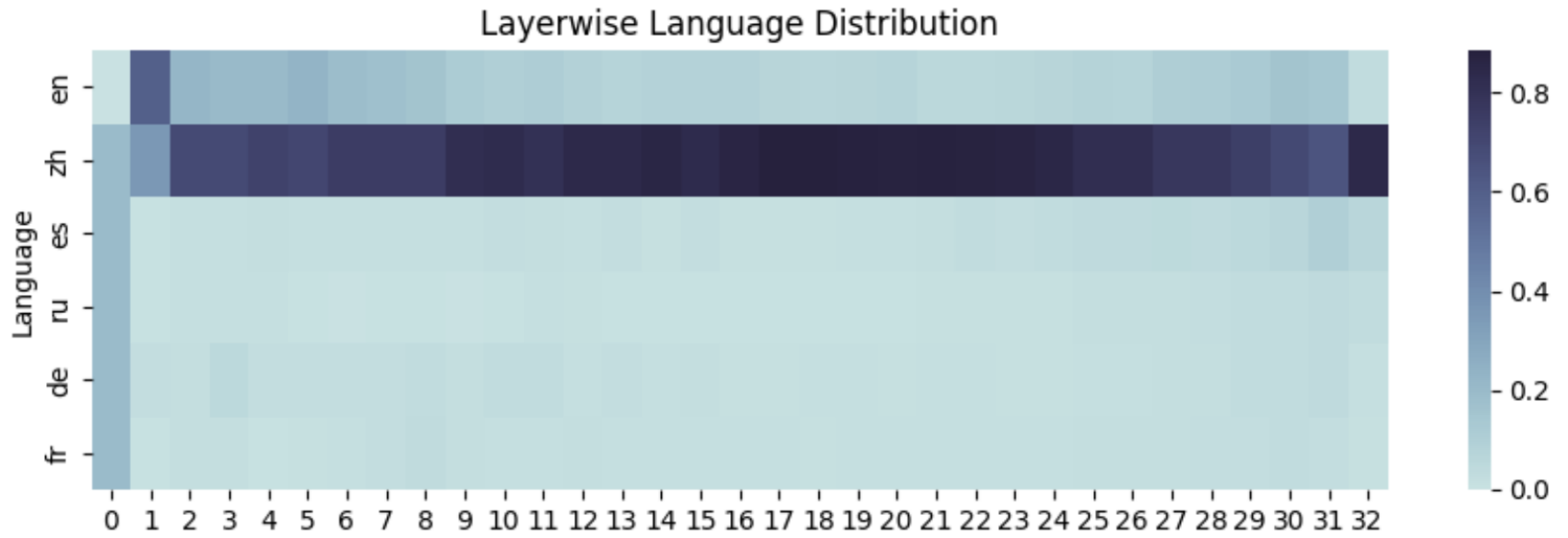}
\caption{Ratio of languages among layers in Chinese Llama given non-English instructions.}
    \label{fig:lang_chatglm}
  \end{minipage}
\vspace{-0.3cm}
\end{figure}

\paragraph{Hyper-Multilingual LLMs}
% From Figure \ref{fig:dis_bloomz} we observe that BLOOMZ also tends to default to English as its ``reasoning'' language. Therfore, it can support multiple languages mainly because it has the ability to understand and generate these languages while using English as the language of reasoning.  
Figure \ref{fig:lang_influen_bloom} illustrates the degree of neuron overlap among languages within both the self-attention and feed-forward structures of BLOOMZ. In contrast to the findings shown in Figure \ref{fig:lang_influ}, there is a marked reduction in overlap, indicating that individual languages maintain a higher degree of independence and do not extensively share neurons with one another.

% \begin{figure}[ht]
% \centering
% \includegraphics[width=\linewidth]{Figures/bloomz_lang.pdf}
%   \caption{Overlapping ratio of language-specific neurons in BLOOMZ.} 
% \label{fig:lang_influen_bloom}
% \vspace{-0.3cm}
% \end{figure}

\paragraph{Bilingual LLMs}
We employ Chinese Llama~\citep{cui2023efficient}, which extends existing vocabulary and incorporate secondary pre-training using Chinese data and fine-tune the model with Chinese instruction datasets. However, this intensive training can lead to a degradation in performance for languages other than Chinese. As depicted in Figure \ref{fig:lang_chatglm}, Chinese predominates as the primary language for reasoning processing and knowledge extraction across all languages. Consequently, the absence of language-specific neurons results in the transformation of it into a Chinese-centric LLM.

% \begin{figure}[ht]
% \centering
% \includegraphics[width=0.5\textwidth]{Figures/chatglm.pdf}
%   \caption{Distribution of languages among layers in Chinese Llama given non-English instructions.} 
% \label{fig:lang_chatglm}
% \end{figure}

\section{Language-Agnostic Neurons}\label{appen_agno}

We initially implement a radical deactivation approach, wherein we specifically deactivate overlapping elements between each language and English. These elements precisely correspond to the intersecting neurons in the first column of Figure \ref{fig:lang_influ}. Presented below are the comprehensive findings pertaining to Mistral. Our evaluation is centered around the reasoning task, which is recognized as the most indicative and challenging assessment for the model. We compare under the optimal ``deactivating'' method, which involves deactivating all language-specific neurons except those in S-ATTN.

\begin{table*}[ht]
\caption{Performance of deactivating language-specific neurons without overlapped between English.}
  \centering
\footnotesize
  \scalebox{0.88}{
  \begin{tabular}{l|c|c|c|c|c}
    \toprule
 \textbf{Language} & Eng & non-Eng & $\Delta_{\text{Eng}}$  & $\Delta_{\text{non-Eng}}$   & $\Delta \uparrow$  \\
    \midrule
 All language-specific neurons & $46.2$ & $18.3$ & $+0.2$ & $-8.0$ & $+8.2$  \\\midrule
 LSN without overlapped between English & $45.8$ & $20.2$ & $-0.2$ & $-6.1$ & $+5.9$    \\
    \bottomrule  
    \end{tabular}}
\label{table:agnostic-non-eng}
\vspace{-0.2cm}
\end{table*}

As evident by Table \ref{table:agnostic-non-eng}, the performance of English remains stable, contrasting sharply with the significant decline in the performance of multilingual. Removing overlapped neurons, as opposed to deactivating all language-specific neurons, leads to a less pronounced drop, yet the impact remains noteworthy. This demonstrates that overlapped neurons are not language-agnostic; they are not utilized for general comprehension and logical reasoning. Otherwise, the fundamental reasoning capacity and performance in multilingual contexts would remain unaffected. In addition, we retained the language-specific neurons that overlapped in all languages, meaning that we removed them from the language-specific neurons to be deactivated. Detailed results follow.

\begin{table*}[ht]
\caption{Performance of deactivating language-specific neurons without all languages overlapped.}
  \centering
\footnotesize
  \scalebox{0.88}{
  \begin{tabular}{l|c|c|c|c|c}
    \toprule
 \textbf{Language} & Eng & non-Eng & $\Delta_{\text{Eng}}$  & $\Delta_{\text{non-Eng}}$   & $\Delta \uparrow$  \\
    \midrule
 All language-specific neurons & $46.2$ & $18.3$ & $+0.2$ & $-8.0$ & $+8.2$  \\\midrule
 LSN without all languages overlapped & $45.6$ & $18.7$ & $-0.4$ & $-7.6$ & $+7.2$    \\
    \bottomrule  
    \end{tabular}}
\label{table:agnostic-all}
\vspace{-0.2cm}
\end{table*}

The neurons that overlap across all languages only account for $0.02\%$ of the total number of neurons. From the results in Table \ref{table:agnostic-all}, we can see that the performance is almost the same as deactivating all language-specific neurons. This further proves that these neurons are not language-agnostic neurons, but only a subset of language-specific neurons.

\section{Prompts}\label{appen:prompt}

Table \ref{table:test_prompt} shows the zero-shot prompts for each dataset. Note that when conducting tests in other languages, prompts are translated into the respective languages.

\begin{table*}[t]
\caption{Zero-shot prompts for each dataset.}  
  \centering
\footnotesize
  \scalebox{1}{
  \begin{tabular}{lp{13cm}}
    \toprule
   \textbf{Task}  & \textbf{Zero-Shot Prompt} \\\midrule
   MGSM & Let's think step by step. Question: \{question\}  \\\midrule
   XQuAD  & \{context\} Question: \{question\} \\ \midrule
   XLSum  & Summarize the context in one sentence. Title: \{title\} Context: \{article\} \\\midrule
   X-CSQA  & Question: \{question\} \\
    \bottomrule
  \end{tabular}}
\label{table:test_prompt}
\end{table*}

\section{Original Performance}\label{appen:orig}

Table \ref{table:vanilla} shows the original performance of Vanilla and Mistral on four tasks.

\begin{table*}[t]
\caption{Assessing the baseline performance of Vicuna and Mistral across four representative multilingual tasks in selected languages, where Avg. is calculated among non-English languages.}
  \centering
\footnotesize
  \scalebox{0.95}{
  \begin{tabular}{l|l|c|c|c|c|c|c|c}
    \toprule
   \normalsize{\textbf{Model}} & \normalsize{\textbf{Task}}  & \normalsize{\textbf{En}} & \normalsize{\textbf{De}} & \normalsize{\textbf{Fr}} & \normalsize{\textbf{Zh}} & \normalsize{\textbf{Es}} & \normalsize{\textbf{Ru}} & \normalsize{\textbf{Avg.}} \\
    \midrule
  \multirow{4}*{{\textbf{Vicuna}}}  & XQuAD & $57.5$ & $50.3$ & $-$ & $55.7$ & $55.7$ & $-$ & $53.9$ \\
  & MGSM & $20.4$ & $14.8$ & $14.8$ & $12.8$ & $13.2$ & $10.0$ & $13.1$ \\
   & X-CSQA & $57.8$ & $43.8$ & $40.1$ & $43.2$ & $44.3$ & $26.0$ & $39.5$ \\
   & XLSum & $13.1$ & $-$ & $14.2$ & $61.1$ & $10.4$ & $20.8$ & $26.6$ \\\midrule
\multirow{4}*{{\textbf{Mistral}}}   & XQuAD & $57.1$ & $48.5$ & $-$ & $64.3$ & $54.1$ & $-$ & $55.6$ \\
  & MGSM & $46.0$ & $21.2$ & $26.0$ & $31.6$ & $31.2$ & $21.6$ & $26.3$ \\
   & X-CSQA & $61.7$ & $40.0$ & $40.4$ & $47.1$ & $45.7$ & $14.1$ & $37.5$\\
   & XLSum & $13.5$ & $-$ & $15.2$ & $56.4$ & $10.6$ & $21.0$ & $25.8$ \\
    \bottomrule  
    \end{tabular}}
\label{table:vanilla}
\end{table*}

\section{Hyper-parameters}\label{appen:hyper}

We adopt the performance on XQuAD in Chinese as the validation set to all languages and all tasks. Specifically, Table \ref{table:result_vicuna} shows the result on Vicuna when deactivating language-specific neurons in the understanding layer (${D}_{\mathcal{U}}$) and generation layer (${D}_{\mathcal{G}}$), where $N_1$ is the number of understanding layers and $N_2$ is the number of generation layer. We find that when setting $N_1 = 8$ and $N_2=2$, performance in English is influenced the least while performance in Chinese decreases the most. As for Mistral, the number is $N_1=6$ and $N_2=3$.

\begin{table*}[ht]
  \centering
  \footnotesize
  \begin{minipage}[b]{0.48\linewidth}
    \centering
\caption{XQuAD with Chinese on Vicuna.}
    \scalebox{0.87}{
      \begin{tabular}{l|cl|cl}
        \toprule
        \multirow{2}*{\textbf{Method}} & \multicolumn{2}{c}{$D_{\mathcal{U}}$} \vline & \multicolumn{2}{c}{$D_{\mathcal{G}}$}    \\
        & $N_1$ &  \multicolumn{1}{c}{$ACC$} \vline & $N_2$ &  \multicolumn{1}{c}{$ACC$} \\\midrule
        En-Vanilla & \multicolumn{4}{c}{$57.5$}    \\
        Zh-Vanilla &  \multicolumn{4}{c}{$55.5$}   \\\midrule
        En-Deact & \multirow{2}*{$\mathbf{8}$} & $\mathbf{57.7}$ (\textcolor[RGB]{84,123,71}{$\uparrow0.2$}) & \multirow{2}*{$4$} & $54.7$ (\textcolor[RGB]{236,89,69}{$\downarrow 2.8$})   \\
        Zh-D-Deact & & $\mathbf{44.9}$ (\textcolor[RGB]{236,89,69}{$\downarrow 10.6$}) & & $54.6$ (\textcolor[RGB]{236,89,69}{$\downarrow 0.9$}) \\\midrule
        En-Deact & \multirow{2}*{$6$} & $58.6$ (\textcolor[RGB]{84,123,71}{$\uparrow1.1$}) & \multirow{2}*{$3$} & $57.7$ (\textcolor[RGB]{84,123,71}{$\uparrow0.2$})  \\
        Zh-Deact & & $55.1$ (\textcolor[RGB]{236,89,69}{$\downarrow 0.4$}) & & $54.5$ (\textcolor[RGB]{236,89,69}{$\downarrow 1.0$})  \\\midrule
        En-Deact  & \multirow{2}*{$4$} & $57.3$ (\textcolor[RGB]{236,89,69}{$\downarrow 0.2$}) & \multirow{2}*{$\mathbf{2}$} & $\mathbf{58.4}$ (\textcolor[RGB]{84,123,71}{$\uparrow0.9$})   \\
        Zh-Deact  & & $53.9$ (\textcolor[RGB]{236,89,69}{$\downarrow 1.6$}) & & $\mathbf{54.1}$ (\textcolor[RGB]{236,89,69}{$\downarrow 1.4$})  \\
        \bottomrule  
      \end{tabular}
    }
    \label{table:result_vicuna}
  \end{minipage}
  \hfill
  \begin{minipage}[b]{0.48\linewidth}
  \caption{XQuAD with Chinese on Mistral.}
    \centering
    \scalebox{0.87}{
      \begin{tabular}{l|cl|cl}
        \toprule
        \multirow{2}*{\textbf{Method}} & \multicolumn{2}{c}{$D_{\mathcal{U}}$} \vline & \multicolumn{2}{c}{$D_{\mathcal{G}}$}    \\
        & $N_1$ &  \multicolumn{1}{c}{$ACC$} \vline & $N_2$ &  \multicolumn{1}{c}{$ACC$} \\\midrule
        En-Vanilla & \multicolumn{4}{c}{$57.1$}    \\
        Zh-Vanilla &  \multicolumn{4}{c}{$64.3$}   \\\midrule
        En-Deact & \multirow{2}*{$8$} & $53.3$ (\textcolor[RGB]{236,89,69}{$\downarrow 3.8$})   & \multirow{2}*{$4$} & $55.8$  (\textcolor[RGB]{236,89,69}{$\downarrow 1.3$})  \\
        Zh-Deact & &  $52.6$ (\textcolor[RGB]{236,89,69}{$\downarrow 11.7$}) & & $62.9$ (\textcolor[RGB]{236,89,69}{$\downarrow 1.4$})  \\\midrule
        En-Deact & \multirow{2}*{$\mathbf{6}$} & $\mathbf{56.8}$ (\textcolor[RGB]{236,89,69}{$\downarrow 0.3$}) & \multirow{2}*{$\mathbf{3}$} & $\mathbf{56.3}$ (\textcolor[RGB]{236,89,69}{$\downarrow 0.8$})  \\
        Zh-Deact & & $\mathbf{54.9}$ (\textcolor[RGB]{236,89,69}{$\downarrow 9.4$}) & & $\mathbf{62.7}$ (\textcolor[RGB]{236,89,69}{$\downarrow 1.6$})   \\\midrule
        En-Deact & \multirow{2}*{$4$} & $57.6$ (\textcolor[RGB]{84,123,71}{$\uparrow0.5$}) & \multirow{2}*{$2$} & $55.7$ (\textcolor[RGB]{236,89,69}{$\downarrow 1.4$})  \\
        Zh-Deact & & $61.8$ (\textcolor[RGB]{236,89,69}{$\downarrow 2.5$}) &  & $63.8$ (\textcolor[RGB]{236,89,69}{$\downarrow 0.5$})  \\
        \bottomrule  
      \end{tabular}
    }
\label{table:result_mistral}
  \end{minipage}
\end{table*}

\section{Detailed Experiment Results}\label{appen:detail}

\subsection{Detailed Experiment Settings}

\paragraph{Reasoning Task} 
Deactivation methods: (i) randomly sampled neurons in the attention structure of task-solving layer. (ii) randomly sampled neurons in the task-solving layer. (iii) randomly sampled neurons in all layers. (iv) language-specific neurons in the task-solving layer. (v) language-specific neurons in the understanding layer and generation layer. (vi) language-specific neurons not in the attention structure of task-solving layers.

\paragraph{Knowledge Question Answering Task}

Deactivation methods: (i) randomly sampled neurons in the feed-forward structure of task-solving layers. (ii) randomly sampled neurons in the task-solving layer. (iii) randomly sampled neurons in all layers. (iv) language-specific neurons in the attention structure of task-solving layers. (v) language-specific neurons in the feed-forward structure of task-solving layers.

\paragraph{Generation Task}

Deactivation methods: (i) randomly sampled neurons in the generating layers. (ii) randomly sampled neurons in all layers. (iv) language-specific neurons in the generating layers.

\subsection{Detailed Result}

Due to the limited space, we employ a more concise notation. We denote deactivating neurons in the self-attention layer of the $i$-th layer as $D^{(A)}_{i}$, while deactivating neurons in the feed-forward layer of the $i$-th layer is denoted as $D^{(F)}_i$.  We denote $\mathcal{U} = \{1,\cdots,N_1\}$ as the set of layers that take charge of understanding as shown in Figure \ref{fig:all}. Similarly, we denote $\mathcal{S} = \{N_1+1,\cdots, N_2\}$ as the set of layers that take charge of task solving and $\mathcal{G} = \{N_2+1,\cdots, 32\}$ as the set of layers that take charge of generation\footnote{{Vicuna-7b-v1.5} and {Mistral-7b-v1.0} both have $32$ layers.}. Furthermore, $D^{(A)}_{\mathcal{U}}$ represents deactivating neurons in self-attention layers of $\mathcal{U}$. Similarly, we introduce $D^{(F)}_{\mathcal{U}}$, $D^{(A)}_{\mathcal{S}}$, $D^{(F)}_{\mathcal{S}}$, $D^{(A)}_{\mathcal{G}}$ and $D^{(A)}_{\mathcal{G}}$.

\begin{table*}[ht]
  \centering
\caption{Understanding task.}
\footnotesize
  \scalebox{0.83}{
  \begin{tabular}{cl|cccc|cccc|cccc}
    \toprule
     & \multirow{2}*{\textbf{\normalsize{Method}}} & \multicolumn{4}{c}{\textbf{\normalsize{German}}} \vline & \multicolumn{4}{c}{\textbf{\normalsize{Chinese}}} \vline & \multicolumn{4}{c}{\textbf{\normalsize{Spanish}}} \\
     & & En-D & De-D & $\Delta_{\text{En-D}}$ & $\Delta_{\text{De-D}}$ & En-D & Zh-D & $\Delta_{\text{En-D}}$ & $\Delta_{\text{Zh-D}}$ & En-D & Es-D & $\Delta_{\text{Es-D}}$ & $\Delta_{\text{Es-D}}$  \\
    \midrule
  \multirow{5}*{\begin{tabular}[c]{@{}l@{}} {\rotatebox{90}{\textbf{\normalsize{Vicuna}}}}
\end{tabular}} 
& $D_{\mathcal{U}}^R$ & \colorbox{white}{$57.8$} & $49.7$ & $+0.3$ & $-0.6$ & $57.8$ & $55.8$ & $+0.3$ & $+0.1$ & $57.8$ & $56.1$ & $+0.3$ & $+0.4$ \\
& $D_{All}^R$ & \colorbox{white}{$57.9$} & $50.8$ & $+0.4$ & $+0.5$ & $57.9$ & $55.8$ & $+0.4$ & $+0.1$ & $57.9$ & $55.9$ & $+0.4$ & $+0.2$ \\
 & $D_{\mathcal{U}}$ & $55.7$ & $40.7$ & \colorbox{lightgray}{$-2.0$} & \colorbox{lightgray}{$-9.6$} & $57.7$ & $44.9$ & \colorbox{lightgray}{$+2.0$} & \colorbox{lightgray}{$-10.8$} & $56.1$ & $52.4$ & \colorbox{lightgray}{$-1.4$} & \colorbox{lightgray}{$-3.2$} \\
 & $D_{\mathcal{S}}$ & \colorbox{white}{$48.3$} & $41.7$ & $-7.2$ & $-8.6$ & $45.0$ & $45.4$ & $-12.5$ & $-10.3$ & $29.5$ & $28.6$ & $-28.0$ & $-27.1$   \\
& $D_{\mathcal{G}}$ & \colorbox{white}{$57.5$} & $50.1$ & $0.0$ & $-0.2$ & $58.4$ & $54.1$ & $+0.9$ & $-1.6$ & $57.7$ & $54.1$ & $+0.2$ & $-1.6$ \\\midrule
  \multirow{5}*{\begin{tabular}[c]{@{}l@{}} {\rotatebox{90}{\textbf{\normalsize{Mistral}}}}
\end{tabular}} 
& $D_{\mathcal{U}}^R$ & \colorbox{white}{$58.1$} & $48.2$ & $+1.0$ & $-0.4$ & $58.1$ & $63.9$ & $+1.0$ & $-0.4$ & $58.1$ & $54.3$ & $+1.0$ & $+0.2$ \\
& $D_{All}^R$ & \colorbox{white}{$57.6$} & $48.3$ & $+0.5$ & $-0.3$ & $57.6$ & $63.6$ & $+0.5$ & $-0.7$ & $57.6$ & $54.5$ & $+0.5$ & $+0.4$ \\
 & $D_{\mathcal{U}}$ & $56.5$ & $42.4$ & \colorbox{lightgray}{$-0.6$} & \colorbox{lightgray}{$-6.2$} & $56.8$ & $54.9$ & \colorbox{lightgray}{$-0.3$} & \colorbox{lightgray}{$-9.4$} & $55.4$ & $47.5$ & \colorbox{lightgray}{$-1.7$} & \colorbox{lightgray}{$-6.6$}  \\
 & $D_{\mathcal{S}}$ & \colorbox{white}{$54.3$} & $43.2$ & $-2.8$ & $-5.4$ & $54.9$ & $52.9$ & $-2.2$ & $-11.4$ & $50.3$ & $44.9$ & $-6.8$ & $-9.2$   \\
& $D_{\mathcal{G}}$ & \colorbox{white}{$56.7$} & $47.9$ & $-0.4$ & $-0.7$ & $56.3$ & $62.7$ & $-0.8$ & $-1.6$ & $56.2$ & $53.2$ & $-0.9$ & $-0.8$ \\
    \bottomrule
    \end{tabular}}
\label{table:result_under}
\end{table*}

\begin{table*}[ht]
  \centering
 \caption{Reasoning task.}
  \footnotesize
  \setlength{\tabcolsep}{1.3pt}
  \scalebox{0.72}{
    \begin{tabular}{cl|cccc|cccc|cccc|cccc|cccc}
      \toprule
      & \multirow{2}{*}{\textbf{Method}} & \multicolumn{4}{c}{\textbf{German}} & \multicolumn{4}{c}{\textbf{French}} & \multicolumn{4}{c}{\textbf{Chinese}} & \multicolumn{4}{c}{\textbf{Spanish}} & \multicolumn{4}{c}{\textbf{Russian}} \\
      & & En-D & De-D & $\Delta_{\text{En-D}}$ & $\Delta_{\text{De-D}}$  & En-D & Fr-D & $\Delta_{\text{En-D}}$ & $\Delta_{\text{Fr-D}}$ & En-D & Zh-D & $\Delta_{\text{En-D}}$ & $\Delta_{\text{Zh-D}}$ & En-D & Es-D & $\Delta_{\text{Es-D}}$ & $\Delta_{\text{Es-D}}$ &  En-D & Ru-D & $\Delta_{\text{En-D}}$ & $\Delta_{\text{Ru-D}}$  \\
      \midrule
      \multirow{6}{*}{\rotatebox{90}{\textbf{Vicuna}}}
      & $D^{R}_{\mathcal{S}^{(A)}}$  & \colorbox{white}{$20.0$}  & $12.4$ & $-0.4$ & $-2.4$ & $20.0$ & $13.6$ & $-0.4$ & $-1.2$ & $20.0$ & $13.2$ & $-0.4$ & $+0.4$ & $20.0$  & $12.4$ & $-0.4$ & $-0.8$ & $20.0$ & $4.8$ & $-0.4$ & $-5.2$ \\
      & $D_{\mathcal{S}}^R$  & \colorbox{white}{$18.4$} & $12.4$ & $-2.0$ & $-2.4$ & $18.4$ & $14.0$ & $-2.0$ & $-0.8$ & $18.4$ & $14.4$ & $-2.0$ & $+1.6$ & $18.4$ & $15.2$ & $-2.0$ & $+2.0$ & $18.4$ & $4.8$ & $-2.0$ & $-5.2$ \\
      & $D_{\text{All}}^R$  & $19.6$ & $14.0$ & $-0.8$ & $-0.8$ & $19.6$ & $13.8$ & $-0.8$ & $-1.0$ & $19.6$ & $14.8$ & $-0.8$ & $+2.0$ & $19.6$ & $12.4$ & $-0.8$ & $-0.8$ & $19.6$ & $7.6$ & $-0.8$ & $-2.4$ \\
     & $D_{\mathcal{S}}$ & $3.6$ & $2.0$ & $-16.8$ & $-12.8$ & $8.4$ & $3.2$ & $-12.0$ & $-11.6$ & $4.8$ & $4.0$ & $-15.6$ & $-8.8$ & $8.8$ & $4.0$ & $-11.6$ & $-9.2$ & $10.4$ & $4.0$ & $-10.0$ & $-6.0$ \\
      & $D_{\mathcal{U}\&\mathcal{G}}$  & \colorbox{white}{$16.4$} & $5.6$ & $-4.0$ & $-9.2$ & $19.2$ & $9.6$ & $-1.2$ & $-5.2$ & $20.0$ & $9.2$ & $-0.4$ & $-3.6$ & $17.6$ & $11.6$ & $-2.8$ & $-1.6$ & $17.2$ & $5.6$ & $-3.2$ & $-4.4$ \\
      & $\bar{D}_{\mathcal{S}^{(A)}}$ & \colorbox{white}{$16.8$} & $4.4$ &  \colorbox{lightgray}{$-3.6$} & \colorbox{lightgray}{$-10.4$} & $19.6$ & $8.8$ &  \colorbox{lightgray}{$-0.8$} &  \colorbox{lightgray}{$-4.4$} & $21.6$ & $9.6$ &  \colorbox{lightgray}{$+1.2$} &  \colorbox{lightgray}{$-3.2$} & $19.6$ & $10.4$ & \colorbox{lightgray}{$-0.8$} & \colorbox{lightgray}{$-2.8$} & $17.2$ & $5.6$ &  \colorbox{lightgray}{$-3.2$} &  \colorbox{lightgray}{$-4.4$} \\
      \midrule
      \multirow{6}{*}{\rotatebox{90}{\textbf{Mistral}}}
       & $D^{R}_{\mathcal{S}^{(A)}}$  & \colorbox{white}{$40.8$} & $18.0$ & $-5.2$ & $-3.2$ & $40.8$ & $25.6$ & $-5.2$ & $-0.4$ & $40.8$ & $24.0$ & $-5.2$ & $-7.6$ & $40.8$ & $29.2$ & $-5.2$ & $-2.0$ & $40.8$ & $20.4$ & $-5.2$ & $-1.2$        \\
      &  $D_{\mathcal{S}}^R$  & \colorbox{white}{$39.2$} & $20.0$ & $-6.8$ & $-1.2$ & $39.2$ & $25.2$ & $-6.8$ & $-0.8$ & $39.2$ & $25.6$ & $-6.8$ &$-6.0$ & $39.2$ & $29.6$ & $-6.8$ & $-1.6$ & $39.2$ & $19.6$ & $-6.8$ & $-2.0$ \\
      & $D_{\text{All}}^R$  & $45.2$ & $24.0$ & $-0.8$ & $+2.8$ & $45.2$ & $27.6$ & $-0.8$ & $+1.6$ & $45.2$ & $31.2$ & $-0.8$ & $-0.4$ & $45.2$ & $30.4$ & $-0.8$ & $-0.8$ & $45.2$ & $20.8$ & $-0.8$ & $-0.8$  \\
      & $D_{\mathcal{S}}$ & \colorbox{white}{$38.4$} & $12.0$ & $-7.6$ & $-9.2$ & $40.8$ & $24.8$ & $-5.2$ & $-1.2$ & $37.9$ & $19.6$ & $-8.1$ & $-12.0$ & $40.4$ & $24.4$ & $-5.6$ & $-6.8$ & $33.6$ & $11.2$ & $-12.4$ & $-10.4$ \\
      & $D_{\mathcal{U}\&\mathcal{G}} $  & \colorbox{white}{$42.4$} & $9.2$ & $-3.6$ & $-12.0$ & $41.2$ & $21.6$ & $-4.8$ & $-4.4$ & $46.4$ & $19.6$ & $+0.4$ & $-12.0$ & $44.0$ & $28.0$ & $-2.0$ & $-3.2$ & $46.0$ & $12.0$ & $+0.0$ & $-9.6$ \\
      & $\bar{D}_{\mathcal{S}^{(A)}}$ & \colorbox{white}{$43.6$} & $9.6$ &  \colorbox{lightgray}{$-2.4$} &  \colorbox{lightgray}{$-11.6$} & $44.8$ & $19.2$ &  \colorbox{lightgray}{$-1.2$} &  \colorbox{lightgray}{$-6.8$} & $46.4$ & $18.8$ &  \colorbox{lightgray}{$+0.4$} &  \colorbox{lightgray}{$-12.8$} & $47.6$ & $27.6$ &  \colorbox{lightgray}{$+1.6$} &  \colorbox{lightgray}{$-3.6$} & $48.4$ & $16.4$ &  \colorbox{lightgray}{$+2.4$} &  \colorbox{lightgray}{$-5.2$} \\
      \bottomrule  
    \end{tabular}}
  \label{table:result_rea}
\end{table*}

\begin{table*}[ht]
  \centering
  \caption{Knowledge Question Answering task.}
  \footnotesize
  \setlength{\tabcolsep}{1.3pt}
  \scalebox{0.72}{
    \begin{tabular}{cl|cccc|cccc|cccc|cccc|cccc}
      \toprule
      & \multirow{2}{*}{\textbf{Method}} & \multicolumn{4}{c}{\textbf{German}} & \multicolumn{4}{c}{\textbf{French}} & \multicolumn{4}{c}{\textbf{Chinese}} & \multicolumn{4}{c}{\textbf{Spanish}} & \multicolumn{4}{c}{\textbf{Russian}} \\
      & & En-D & De-D & $\Delta_{\text{En-D}}$ & $\Delta_{\text{De-D}}$  & En-D & Fr-D & $\Delta_{\text{En-D}}$ & $\Delta_{\text{Fr-D}}$ & En-D & Zh-D & $\Delta_{\text{En-D}}$ & $\Delta_{\text{Zh-D}}$ & En-D & Es-D & $\Delta_{\text{Es-D}}$ & $\Delta_{\text{Es-D}}$ &  En-D & Ru-D & $\Delta_{\text{En-D}}$ & $\Delta_{\text{Ru-D}}$  \\
      \midrule
      \multirow{5}{*}{\rotatebox{90}{\textbf{Vicuna}}}
      & $D^{R}_{\mathcal{S}^{(F)}}$  &  \colorbox{white}{$57.5$} & $43.8$ & $-0.3$ & $+0.0$ & $57.5$ & $40.3$ & $-0.3$ & $+0.2$ & $57.5$ & $43.2$ & $-0.3$ & $+0.0$ & $57.5$ & $44.6$ & $-0.3$ & $+0.3$ & $57.5$ & $25.5$ & $-0.3$ & $-0.5$ \\
      & $D_{\mathcal{S}}^R$  &  \colorbox{white}{$56.0$} & $44.0$ & $-1.8$ & $+0.2$ & $56.0$ & $38.6$ & $-1.8$ & $-1.5$ & $56.0$ & $43.4$ & $-1.8$ & $+0.2$ & $56.0$ & $43.5$ & $-1.8$ & $-0.8$ & $56.0$ & $24.0$ & $-1.8$ & $-2.0$ \\
      & $D_{\text{All}}^R$  &  \colorbox{white}{$57.7$} & $43.6$ & $-0.1$ & $-0.2$ & $57.7$ & $40.5$ & $-0.1$ & $+0.4$ & $57.7$ & $43.2$ & $-0.1$ & $+0.0$ & $57.7$ & $44.5$ & $-0.1$ & $+0.2$ & $57.7$ & $26.0$ & $-0.1$ & $+0.0$ \\
      & $D_{\mathcal{S}^{(A)}}$ & \colorbox{white}{$34.8$} & $43.4$ & $-23.0$ & $-0.4$ & $32.6$ & $31.1$ & $-25.2$ & $-12.7$ & $32.6$ & $28.9$ & $-25.2$ & $-14.3$ & $20.4$ & $25.0$ & $-37.1$ & $-19.3$ & $48.3$ & $22.9$ & $-9.5$ & $-3.1$  \\
      & ${D}_{\mathcal{S}^{(F)}}$ & $57.8$ & $41.5$ & \colorbox{lightgray}{$+0.0$} & \colorbox{lightgray}{$-2.5$} & $57.2$ & $37.8$ & \colorbox{lightgray}{$-0.6$} & \colorbox{lightgray}{$-6.0$} & $56.9$ & $39.6$ & \colorbox{lightgray}{$-0.9$} & \colorbox{lightgray}{$-3.6$} & $57.6$ & $43.0$ & \colorbox{lightgray}{$-0.2$} & \colorbox{lightgray}{$-1.3$} & $57.8$ & $25.6$ & \colorbox{lightgray}{$+0.0$} & \colorbox{lightgray}{$-0.4$}  \\
      \midrule
      \multirow{5}{*}{\rotatebox{90}{\textbf{Mistral}}}
      & $D^{R}_{\mathcal{S}^{(F)}}$ & \colorbox{white}{$61.0$} & $40.2$ & $-0.7$ & $+0.2$ & $61.0$ & $40.1$ & $-0.7$ & $-0.3$ & $61.0$ & $46.7$ & $-0.7$ & $-0.4$ & $61.0$ & $45.2$ & $-0.7$ & $-0.5$ & $61.0$ & $12.7$ & $-0.7$ & $-1.4$  \\
      & $D_{\mathcal{S}}^R$  & \colorbox{white}{$60.7$} & $40.4$ & $-1.0$ & $+0.4$ & $60.7$ & $36.9$ & $-1.0$ & $-3.5$ & $60.7$ & $46.9$ & $-1.0$ & $-0.3$ & $60.7$ & $46.3$ & $-1.0$ & $+0.7$ & $60.7$ & $11.1$ & $-1.0$ & $-3.0$ \\
      & $D_{\text{All}}^R$  & \colorbox{white}{$61.8$} & $40.1$ & $+0.1$ & $+0.1$ & $61.8$ & $40.7$ & $+0.1$ & $+0.3$ & $61.8$ & $47.2$ & $+0.1$ & $+0.1$ & $61.8$ & $44.7$ & $+0.1$ & $-1.0$ & $61.8$ & $14.1$ & $+0.1$ & $+0.0$ \\
      & $D_{\mathcal{S}^{(A)}}$ & \colorbox{white}{$50.4$} & $32.3$ & $-11.3$ & $-7.7$ & $55.3$ & $27.4$ & $-6.4$ & $-13.0$ & $54.7$ & $42.4$ & $-7.0$ & $-4.7$ & $44.5$ & $34.1$ & $-17.2$ & $-11.6$ & $51.1$ & $8.3$ & $-10.6$ & $-5.8$ \\
      & ${D}_{\mathcal{S}^{(F)}}$ & $61.5$ & $38.1$ & \colorbox{lightgray}{$-0.2$} & \colorbox{lightgray}{$-1.9$} & $61.2$ & $38.1$ & \colorbox{lightgray}{$-0.5$} & \colorbox{lightgray}{$-2.3$} & $61.3$ & $43.5$ & \colorbox{lightgray}{$-0.4$} & \colorbox{lightgray}{$-3.6$} & $61.0$ & $43.9$ & \colorbox{lightgray}{$-0.7$} & \colorbox{lightgray}{$-1.8$} & $60.8$ & $11.8$ & \colorbox{lightgray}{$-0.4$} & \colorbox{lightgray}{$-2.3$}  \\
      \bottomrule  
    \end{tabular}}
  \label{table:result_know}
\end{table*}

\begin{table*}[ht]
  \centering
\caption{Generation task.}
\footnotesize
  \setlength{\tabcolsep}{3pt}
  \scalebox{0.8}{
  \begin{tabular}{cl|cccc|cccc|cccc|cccc}
    \toprule
     & \multirow{2}*{\textbf{\normalsize{Method}}} & \multicolumn{4}{c}{\textbf{\normalsize{French}}} \vline & \multicolumn{4}{c}{\textbf{\normalsize{Chinese}}} \vline & \multicolumn{4}{c}{\textbf{\normalsize{Spanish}}} \vline & \multicolumn{4}{c}{\textbf{\normalsize{Russian}}} \\
     & & En-D & Fr-D & $\Delta_{\text{En-D}}$ & $\Delta_{\text{Fr-D}}$ & En-D & Zh-D & $\Delta_{\text{En-D}}$ & $\Delta_{\text{Zh-D}}$ & En-D & Es-D & $\Delta_{\text{Es-D}}$ & $\Delta_{\text{Es-D}}$ &  En-D & Ru-D & $\Delta_{\text{En-D}}$ & $\Delta_{\text{Ru-D}}$  \\
    \midrule
  \multirow{3}*{\begin{tabular}[c]{@{}l@{}} {\rotatebox{90}{\textbf{{Vicuna}}}}
\end{tabular}} 
& $D_{\mathcal{G}}^R$  & \colorbox{white}{$13.2$} & $14.2$ & $+0.1$ & $+0.0$ & $13.2$ & $61.6$ & $+0.1$ & $+0.5$ & $13.2$ & $10.4$ & $+0.1$ & $+0.0$ & $13.2$ & $20.8$ & $+0.1$ & $+0.0$  \\
& $D_{All}^R$ & $13.0$ & $14.1$ & $-0.1$ & $-0.1$ & $13.0$ & $61.6$ & $-0.1$ & $+0.5$ & $13.0$ & $10.4$ & $-0.1$ & $+0.0$ & $13.0$ & $20.8$ & $-1.0$ & $+0.0$\\
& $D_{\mathcal{G}}$  & \colorbox{white}{$13.0$} & $13.8$ & \colorbox{lightgray}{$-0.1$} & \colorbox{lightgray}{$-0.4$} & $13.1$ & $59.5$ & \colorbox{lightgray}{$+0.0$} & \colorbox{lightgray}{$-1.6$} & $13.0$ & $9.1$ & \colorbox{lightgray}{$-0.1$} & \colorbox{lightgray}{$-1.3$} & $13.1$ & $20.3$ & \colorbox{lightgray}{$+0.0$} & \colorbox{lightgray}{$-0.5$} \\\midrule
  \multirow{3}*{\begin{tabular}[c]{@{}l@{}} {\rotatebox{90}{\textbf{{Mistral}}}}
\end{tabular}} 
& $D_{\mathcal{G}}^R$   &  \colorbox{white}{$13.6$} & $15.2$ & $+0.1$ & $+0.0$ & $13.6$ & $56.7$ & $+0.1$ & $+0.3$ & $13.6$ & $10.3$ & $+0.1$ & $-0.3$ & $13.6$ & $21.2$ & $+0.1$ & $+0.2$  \\
& $D_{All}^R$  & $13.6$ & $15.4$ & $+0.1$ & $+0.2$ & $13.6$ & $55.9$ & $+0.1$ & $-0.5$ & $13.6$ & $10.2$ & $+0.1$ & $-0.4$ & $13.6$ & $21.1$ & $+0.1$ & $+0.1$ \\
& $D_{\mathcal{G}}$  & $14.3$ & $14.2$ & \colorbox{lightgray}{$+0.8$} &  \colorbox{lightgray}{$-1.0$} & $13.6$ & $52.8$ &  \colorbox{lightgray}{$+0.1$} &  \colorbox{lightgray}{$-3.6$} & $13.7$ & $10.2$ &  \colorbox{lightgray}{$+0.2$} &  \colorbox{lightgray}{$-0.4$} & $13.5$ & $20.2$ &  \colorbox{lightgray}{$-0.1$} &  \colorbox{lightgray}{$-0.8$}  \\
    \bottomrule  
    \end{tabular}}
    \label{table:result_gen}
\end{table*}

\end{document}